\documentclass{article}

\usepackage[preprint,nonatbib]{neurips_2021}

\usepackage{epsfig}
\usepackage{graphicx}
\usepackage{amsmath}

\usepackage{enumerate}

\usepackage[font=small,labelfont=bf]{caption}

\usepackage{xcolor}
\usepackage{amsthm}
\usepackage{amsfonts}
\usepackage[vlined, ruled]{algorithm2e}
\usepackage{setspace}
\usepackage{caption}
\usepackage{subcaption}
\usepackage{bm}
\usepackage{isomath}
\usepackage{algorithm2e}

\usepackage[pagebackref=true,breaklinks=true,letterpaper=true,colorlinks,bookmarks=false]{hyperref}

\usepackage{amssymb}

\usepackage{stmaryrd}
\usepackage{float}
\usepackage{pbox}
\usepackage{color}
\usepackage[normalem]{ulem}
\usepackage[toc,page]{appendix}
\usepackage{cleveref}
\usepackage{url}
\usepackage{enumitem}
\usepackage{arydshln}
\usepackage{multirow}
\usepackage{bbding}

\usepackage{booktabs}

\renewcommand\vec[1]{\ensuremath\boldsymbol{#1}}
\renewcommand\cdots{...}

\newcommand{\vb}{\mathbf{b}}
\newcommand{\vy}{\mathbf{y}}

\newcommand{\tX}{\vec{\mathcal{X}}}

\newcommand{\vx}{\mathbf{x}}

\newcommand{\mbr}[1]{\mathbb{R}^{#1}}

\newcommand{\mvv}{\mathbf{v}}

\newcommand{\idx}[1]{\mathcal{I}_{#1}}

\newcommand{\vu}{\mathbf{u}}

\newcommand{\vzeta}{\boldsymbol{\zeta}}

\newcommand{\vphi}{\boldsymbol{\phi}}
\newcommand{\vpsi}{\boldsymbol{\psi}}

\newcommand{\vupsilon}{\boldsymbol{\upsilon}}
\newcommand{\vdelta}{\boldsymbol{\delta}}
\newcommand{\mupsilon}{\boldsymbol{\Upsilon}}

\newcommand{\iu}{{i\mkern1mu}}

\newcommand{\enorm}[1]{\left\|{#1}\right\|_2}

\DeclareMathOperator*{\kronstack}{\uparrow\!\otimes}

\newcommand{\expl}[1]{\text{e}^{#1}}

\newcommand{\mI}{\mathbf{I}}

\newtheorem{proposition}{Proposition}
\newtheorem{remark}{Remark}

\newcommand{\mLambda}{\bm{\lambda}}
\newcommand{\mU}{\bm{U}}
\newcommand{\mV}{\bm{V}}

\newcommand{\mTheta}{\bm{\theta}}
\newcommand{\mLLa}{\bm{\Lambda}}

\newcommand{\vsss}{\boldsymbol{s}}
\newcommand{\vh}{\boldsymbol{h}}

\def\eg{\emph{e.g.}}

\newcommand{\mM}{\boldsymbol{M}}

\newcommand{\mW}{\boldsymbol{W}}

\newcommand{\vkappa}{\boldsymbol{\kappa}}
\newcommand{\vmu}{\boldsymbol{\mu}}

\newcommand{\mP}{\boldsymbol{\Theta}}
\newcommand{\mPP}{\boldsymbol{P}}

\newcommand{\stkout}[1]{{\ifmmode\text{\sout{\ensuremath{#1}}}\else\sout{#1}\fi}}

\makeatletter
\DeclareRobustCommand\onedot{\futurelet\@let@token\bmv@onedotaux}
\def\bmv@onedotaux{\ifx\@let@token.\else.\null\fi\xspace}
\def\eg{\emph{e.g}\onedot} 
\def\ie{\emph{i.e}\onedot} 
\def\cf{\emph{c.f}\onedot} 
\def\etc{\emph{etc}\onedot} \def\vs{\emph{vs}\onedot}
\def\wrt{w.r.t\onedot}

\makeatother

\title{Self-supervising Action Recognition by Statistical Moment and Subspace Descriptors} 

\author{%
  Lei Wang, Piotr Koniusz
	\thanks{The corresponding author. This paper is published at the ACM MM'21. DOI: \url{https://doi.org/10.1145/3474085.3475572}. The code will be available at \url{http://users.cecs.anu.edu.au/\~koniusz/}} 
	\\
  Australian National University and Data61/CSIRO\\
  Canberra, Australia \\
  \texttt{\{lei.w,piotr.koniusz\}@anu.edu.au} \\
}

\begin{document}

\maketitle

\begin{abstract}
In this paper,  we build on a concept of self-supervision by taking RGB frames as input to learn to predict both action concepts and auxiliary  descriptors \eg, 
object descriptors. 
So-called hallucination streams are trained to predict auxiliary cues, simultaneously fed into classification layers, and then  hallucinated at the testing stage to aid network. 
%
We design and hallucinate two descriptors, one leveraging four popular object detectors applied to training videos, and the other leveraging image- and video-level saliency detectors. 
The first descriptor encodes the detector- and ImageNet-wise class prediction scores, confidence scores, and spatial locations of bounding boxes and frame indexes to capture the spatio-temporal distribution of features per video. Another descriptor encodes spatio-angular gradient distributions of saliency maps and intensity patterns. 
Inspired by the characteristic function of the probability distribution, we capture four statistical moments on the above intermediate descriptors. As numbers of coefficients in the  mean, covariance, coskewness and cokurtotsis grow linearly, quadratically, cubically and quartically \wrt the dimension of feature vectors, we describe the covariance matrix by its leading $n'\!$ eigenvectors (so-called subspace) and we capture skewness/kurtosis rather than costly coskewness/cokurtosis. We obtain state of the art on  five popular datasets such as Charades and EPIC-Kitchens.

\end{abstract}


\section{Introduction}
\label{sec:intro}

\begin{figure}[t]
\centering

%
\begin{subfigure}[b]{0.495\linewidth}
\centering\includegraphics[trim=0 0 0 0, clip=true,width=0.95\linewidth]{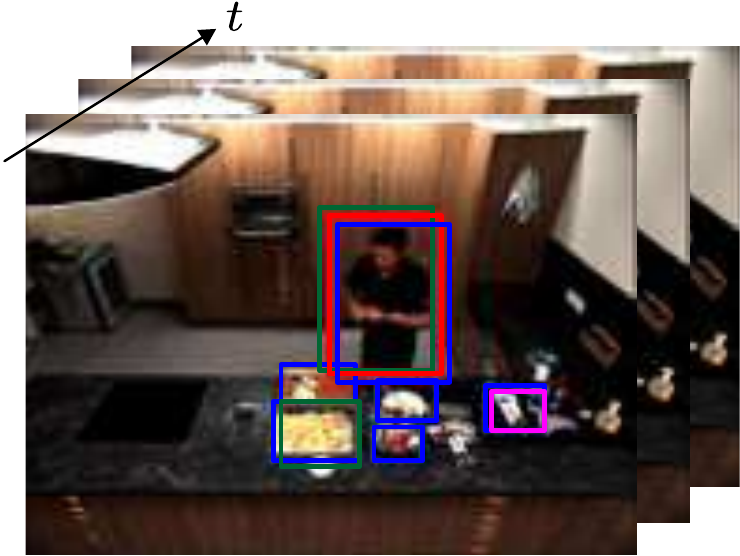}
\vspace{-0.2cm}
\caption{\label{fig:det}\vspace{-0.2cm}}
\end{subfigure}
\begin{subfigure}[b]{0.495\linewidth}
\begin{subfigure}[b]{0.995\linewidth}
\centering\includegraphics[trim=0 0 0 0, clip=true,width=0.90\linewidth]{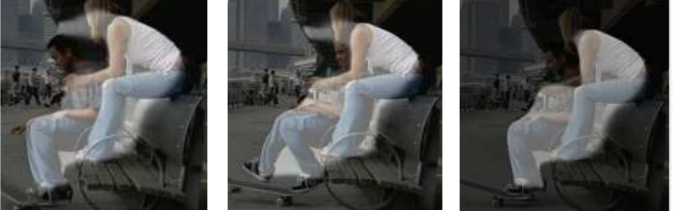}
\vspace{-0.2cm}
\caption{\label{fig:sal-reg}}
\end{subfigure}
\begin{subfigure}[b]{0.995\linewidth}
\centering\includegraphics[trim=0 0 0 0, clip=true,width=0.90\linewidth]{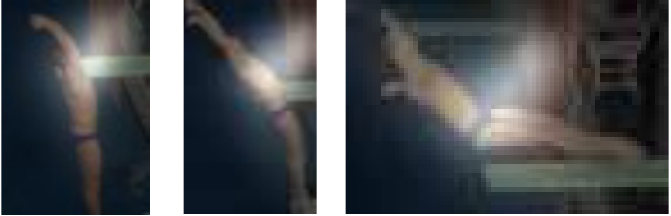}
\vspace{-0.2cm}
\caption{\label{fig:sal-temp}}
\end{subfigure}
\end{subfigure}
\caption{We use detectors/saliency in hallucination descriptors. Figure \ref{fig:det} shows bounding boxes from four detectors. The faster R-CNN detector with ResNet101 focuses on human-centric actions such as {\em stand}, {\em watch}, {\em talk}, \etc. The other three detectors discover  \eg, {\em oven}, {\em sink}, {\em clock}, \etc. Figure \ref{fig:sal-reg} shows that the MNL saliency detector focuses on spatial regions. Figure \ref{fig:sal-temp} shows ACLNet saliency detector discovers motion regions.}
\vspace{-0.3cm}
\label{fig:det-sal}
\end{figure}

Action Recognition (AR) has progressed from hand-crafted video representations \cite{hof,sift_3d,3D-HOG,dense_traj,dense_mot_boundary,improved_traj,lei_thesis_2017,lei_tip_2019} to  Convolutional Neural Networks (CNN) \cite{two_stream,spattemp_filters,spat_temp_resnet,i3d_net}. The two-stream networks \cite{two_stream}, 3D spatio-temporal features \cite{spattemp_filters}, spatio-temporal ResNet model \cite{spat_temp_resnet} and the new Inflated 3D (I3D) convolutions network pre-trained on Kinetics-400 \cite{i3d_net}. Often, AR combine the RGB and optical flow inputs, and  benefit from a late fusion (next to the classifier) with low-level representations such as Improved Dense Trajectory (IDT) descriptors \cite{improved_traj} due to their highly complementary nature \cite{basura_rankpool2,hok,anoop_rankpool_nonlin,anoop_advers,potion}. 
Recently, AssembleNet and AssembleNet++ \cite{assemblenet}, learnt with the Neural Architecture Search (NAS) have yielded superb results.

A recent AR pipeline \cite{Wang_2019_ICCV}, called DEEP-HAL, used IDT descriptors encoded with Bag-of-Words (BoW) \cite{sivic_vq,csurka04_bovw} and Fisher Vectors (FV) \cite{perronnin_fisher,perronnin_fisherimpr} to learn them by so-called hallucination streams and generate at the testing stage to boost results beyond a naive fusion of modalities. DEEP-HAL and approach \cite{tencent_hall} have shown that even optical flow frames encoded by a network can be learnt by another network trained on RGB frames only, thus pointing at redundancy in training both RGB and optical flow network streams. DEEP-HAL \cite{Wang_2019_ICCV} has attained the state of the art  on several AR benchmarks by learning to hallucinate IDT-based BoW/FV and Optical Flow Features (OFF) from a single RGB-based I3D network stream.

DEEP-HAL opens up an exciting opportunity to investigate what other representations can co-regularize/self-supervise a backbone network for AR with the goal of learning to hallucinate costly representations at the training stage and simply leveraging outputs of halluciantion streams at the testing time. 
We build on DEEP-HAL which already includes IDT-based BoW/FV and OFF streams. However, we investigate the self-supervisory ability of object/saliency detectors in DEEP-HAL. Moreover, beyond I3D backbone, we investigate the use of AssembleNet and AssembleNet++ but we disable their (impractical to obtain) segmentation mask input.

In this paper, we design and hallucinate two kinds of descriptors, namely Object Detection Features (ODF) and Saliency Detection Features (SDF). The ODF descriptor leverages faster R-CNN detector \cite{faster-rcnn} based on backbones such as (i) Inception V2~\cite{Szegedy_2016_CVPR}, (ii) Inception ResNet V2~\cite{Szegedy_2017_AAAI}, (iii) ResNet101~\cite{He_2016_CVPR} and (iv) NASNet~\cite{Zoph_2018_CVPR}. The Inception V2, Inception ResNet V2 and NASNet are pre-trained on the COCO dataset~\cite{Lin_eccv2014_coco} (91 object classes), whereas the ResNet101 is pre-trained on the AVA v2.1 dataset~\cite{Gu_2018_CVPR} (80 human AR classes). The above detectors are applied to training videos to identify humans and objects. Such detected objects  together with their relevance and class labels summarized with our descriptor encourage the AR pipeline to focus on  semantically important regions and actors  relevant to the task of action recognition. Figure \ref{fig:det} shows a few of bounding boxes detected by these four detectors.

\begin{figure*}[b]
\vspace{-0.3cm}
\centering\includegraphics[trim=0 0 0 0, clip=true,width=0.95\linewidth]{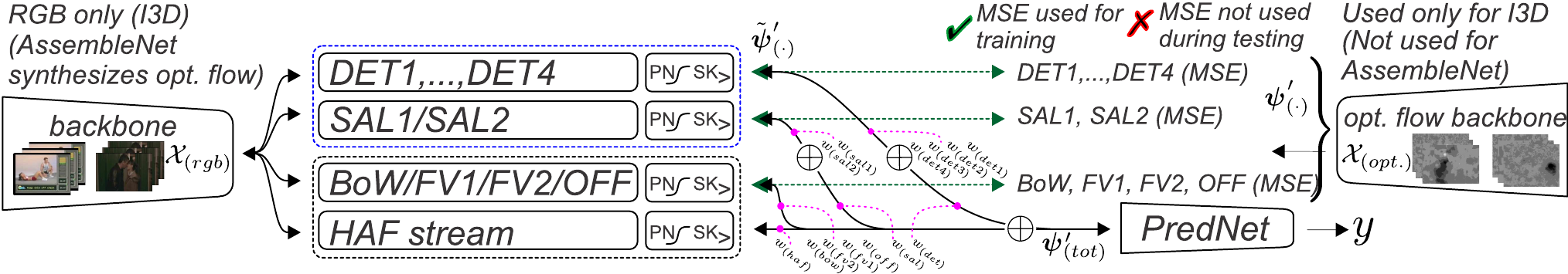}
\caption{We build on DEEP-HAL \cite{Wang_2019_ICCV} which includes I3D RGB and Optical Flow networks (the latter net. is used only during training). For AssembleNet and AssembleNet++, the backbone encodes both RGB and the optical flow, which is synthesized on the fly from RGB frames. 
For the I3D variant, we remove the prediction and the last 1D conv. layers from I3D RGB and optical flow streams, we feed the $1024\!\times\!7$ feature representations $\mathcal{X}_{(rgb)}$ into {\em Bag-of-Words} ({\em BoW}), {\em Fisher Vector} ({\em FV}), {\em the Optical Flow Features} ({\em OFF}) and {\em the High Abstraction Features} ({\em HAF}) streams (dashed black) followed by the {\em Power Normalization} ({\em PN}) and {\em Sketching} ({\em SK}) blocks. 
The OFF stream is supervised by $\mathcal{X}_{(opt.)}$. For the AssembleNet variant, we obtain the $2048$ feature representations $\mathcal{X}_{(rgb)}$ and do not use the OFF stream/optical flow backbone. 
Moreover, we introduce {\em DET1}$,\cdots,${\em DET4}, {\em SAL1} and {\em SAL2} streams corresponding to our detector- and saliency-based descriptors (dashed blue). The resulting feature vectors $\vec{\tilde{\psi}}'_{(\cdot)}$, where $(\cdot)$ denotes the stream name \eg, $(det1),\cdots,(det2)$ \etc, are reweighted by corresponding weights $w_{(\cdot)}$ (magenta lines) and aggregated (sum) by ($\oplus$). 
All $\vec{\tilde{\psi}}_{(\cdot)}$ are reweighted,  aggregated (sum) and fed to {\em Prediction Network} ({\em PredNet}). 
By \Checkmark, we indicate that the Mean Square Error (MSE) losses are used during training 
to supervise all streams outputting $\vec{\tilde{\psi}}'_{(\cdot)}$ by the  ground-truth  $\vpsi'_{(\cdot)}$. By \XSolidBrush, we indicate that the MSE losses are switched off for testing and $\vec{\tilde{\psi}}'_{(\cdot)}$ are hallucinated/fed into PredNet to obtain labels $y$. 
}
\vspace{-0.3cm}
\label{fig:pipe}
\end{figure*}

The SDF  leverages image- and video-level saliency detectors such as MNL~\cite{Zhang_2018_CVPR} and ACLNet~\cite{Zhang_2019_CVPR} with the goal of identifying salient regions correlating with the human gaze in spatial and temporal sense. Saliency maps extracted from training videos and summarized by our descriptor help the AR pipeline learn spatial and temporal regions correlating with actions. Figures \ref{fig:sal-reg} and \ref{fig:sal-temp} show saliency maps from region-wise and temporal saliency detectors.

IDT descriptors are fused with the majority of  modern AR pipelines \cite{basura_rankpool2,hok,anoop_rankpool_nonlin,anoop_advers,potion} at the classifier level for the best performance while DEEP-HAL \cite{Wang_2019_ICCV} learns to hallucinate, and feeds them into the classification branch called PredNet. In this paper, we go further and prepare two  compact descriptors, ODF and SDF, and hallucinate them within DEEP-HAL. We equip each hallucination branch with a weighting mechanism adjusted per epoch to attain the best results. Figure \ref{fig:pipe} illustrates DEEP-HAL at the conceptual level.

For ODF descriptors, we concatenate together per bounding box per frame (i) the one-hot detection and (ii) ImageNet \cite{ILSVRC15} scores, (iii) embedded confidence scores, (iv) embedded bounding box coordinates, and (v) embedded normalized frame index. For all bounding boxes, we stack such features into a matrix. Inspired by the characteristic function of the probability density fun., we extract the mean, leading eigenvectors of covariance, skewness and kurtosis. 
For SDF descriptors, per frame, we encode saliency via (i) kernelized descriptor on spatio-angular gradient distributions of saliency maps and (ii) intensity patterns. %
We obtain an ODF per detector and an SDF per saliency detector. Our contributions are as follows:
\renewcommand{\labelenumi}{\roman{enumi}.}
/\vspace{-0.1cm}
\hspace{-1cm}
\begin{enumerate}[leftmargin=0.6cm]
\item We propose to use the object and human detectors to enhance the performance of AR pipelines.
\item We design two types of statistically motivated high-order compact descriptors, Object Detection Features and Saliency Detection Features, for the use in AR pipelines. 
\item We build on the recent DEEP-HAL  pipeline \cite{Wang_2019_ICCV} but we introduce AssembleNet and AssembleNet++ apart from I3D backbone. Moreover, we introduce a weight learning mechanism for hallucinated feature vectors, and  ODF and SDF are hallucinated which leads to the state-of-the-art performance.
\end{enumerate}



\section{Related Work}
\label{sec:related}



Below, we describe handcrafted spatio-temporal video descriptors, their encoding strategies and the optical flow used by DEEP-HAL \cite{Wang_2019_ICCV}. We also describe deep learning pipelines for video classification. Finally, we discuss the object category and human detectors followed by the spatial and temporal saliency detectors used by us. 

\vspace{0.05cm}
\noindent{\bf{Early video descriptors.}} Early AR 
used on spatio-temporal interest point detectors \cite{harris3d,cuboid,sstip,hes-stip,mv-stip,dense_traj} and spatio-temporal descriptors \cite{hof,sift_3d,hof2, dense_traj,dense_mot_boundary,improved_traj} which capture various appearance and motion statistics. 
%
%
As spatio-temporal interest point detectors are unable to capture long-term motion patterns, a Dense Trajectory (DT) \cite{dense_traj} approach densely samples feature points in each frame to track them in the video (via optical flow). Then, multiple descriptors are extracted along trajectories to capture shape, appearance and motion cues. As DT cannot compensate for the camera motion, the IDT \cite{improved_traj,dense_mot_boundary}  estimates the camera motion to remove the global background motion. IDT also removes inconsistent matches via a human detector. 
For spatio-temporal descriptors, IDT employs HOG \cite{hog2d}, HOF \cite{hof} and MBH \cite{dense_mot_boundary}. 
HOG \cite{hog2d} contains statistics of the amplitude of image gradients \wrt the gradient orientation, thus it captures the static appearance cues. 
In contrast, HOF \cite{hof}  captures histograms of optical flow while MBH \cite{dense_mot_boundary} captures derivatives of the optical flow, thus it is highly resilient to the global camera motion whose cues cancel out due to derivatives. Thus, HOF and MBH contain the zero- and first-order optical flow statistics. Other spatio-temporal descriptors include HOG-3D \cite{3D-HOG}, SIFT3D \cite{sift_3d}, SURF3D \cite{hes-stip} and 
LTP \cite{LTP}.

We use the  DEEP-HAL \cite{Wang_2019_ICCV} setup. We encode  HOG, HOF, and MBH descriptors on the Improved Dense Trajectories  \cite{dense_traj,hok,potion} via BoW \cite{sivic_vq,csurka04_bovw} and FV \cite{perronnin_fisher,perronnin_fisherimpr}.

\vspace{0.05cm}
\noindent{\bf{BoW/FV encoding.}} BoW \cite{sivic_vq,csurka04_bovw} 
uses a k-means vocabulary to which local descriptors are assigned. Variants include Soft Assignment (SA) \cite{soft_ass,me_SAO} and Localized Soft Assignment (LcSA) \cite{liu_sadefense,me_ATN}. 
%
As we use  DEEP-HAL \cite{Wang_2019_ICCV}, we use BoW  \cite{csurka04_bovw} with Power Normalization \cite{me_ATN}, and FV \cite{perronnin_fisher,perronnin_fisherimpr} which capture first- and second-order statistics of local descriptors assigned to GMM clusters. 
DEEP-HAL \cite{Wang_2019_ICCV} setup  describes how to obtain the BoW/FV global  descriptors.

\vspace{0.05cm}
\noindent{\bf{Optical flow.}}  
Older optical flow methods cope with small displacements \cite{flow_def2,brox_accurate} while newer methods cope with larger displacements \eg, Large Displacement Optical Flow (LDOF) \cite{brox_largedisp}. Recent methods use non-rigid descriptor or segment matching \cite{deep_flow,seg_flow}, or edge-preserving interpolation \cite{epic_flow}. 
We use LDOF \cite{brox_accurate}.

\vspace{0.05cm}
\noindent{\bf{Object detectors.}}  
Modern deep learning methods include Region-based Convolutional Neural Networks (R-CNN) \cite{rcnn}, its faster variants \cite{fast_rcnn,faster-rcnn}, its mask-based variants \cite{mask_rnn}, and YOLO \cite{yolo}, YOLO v2, YOLO v3 \etc, which use a single network for efficiency.

In this paper, we use the faster R-CNN detector \cite{faster-rcnn} with backbones such as (i) Inception V2~\cite{Szegedy_2016_CVPR}, (ii) Inception ResNet V2~\cite{Szegedy_2017_AAAI}, (iii) ResNet101~\cite{He_2016_CVPR} and (iv) NASNet~\cite{Zoph_2018_CVPR}. As the Inception V2, Inception ResNet V2 and NASNet are pre-trained on the COCO dataset~\cite{Lin_eccv2014_coco}, they detect from 91 object classes good at summarizing \eg, indoor environments and helping us associate the scene context with actions. The ResNet101 model is pre-trained on the AVA v2.1 dataset~\cite{Gu_2018_CVPR} with 80 different human actions, thus directly helping human-centric action recognition problems.

In addition to detection scores, we describe each bounding box with ImageNet \cite{ILSVRC15} scores from pre-trained Inception ResNet V2~\cite{Szegedy_2017_AAAI}.

\vspace{0.05cm}
\noindent{\bf{Saliency detectors.}} 
Image regions  correlating with human visual attention are detected by saliency detectors in the form of saliency maps. 
Deep saliency models \cite{RFCN,ChengCVPR17} outperform  conventional saliency detectors \cite{Background-Detection:CVPR-2014} but they require  pixel-wise annotations. Recent  models include MNL \cite{Zhang_2018_CVPR} (weakly-supervised model), RFCN \cite{RFCN} (a fully-supervised model) and  a cheap non-CNN Robust Background Detector (RBD) \cite{Background-Detection:CVPR-2014} (see survey \cite{SalObjBenchmark_Tip2015} for more details).

For the spatial  saliency, we use MNL \cite{Zhang_2018_CVPR} 
trained on multiple noisy labels from weak/noisy unsupervised handcrafted saliency models. 
For temporal saliency, we use a CNN-LSTM  ACLNet \cite{Zhang_2019_CVPR}. 

\vspace{0.05cm}
\noindent{\bf{Deep learning AR.}} 
Early AR CNN models use frame-wise features and average pooling 
\cite{cnn_basic_ar} discarding the temporal order. Thus, frame-wise CNN scores were fed to LSTMs \cite{cnn_lstm_ar} while the two-stream networks \cite{two_stream} compute representations per RGB frame and per 10 stacked optical flow frames. Finally, spatio-temporal 3D CNN filters \cite{cnn3d_ar,spattemp_filters,spat_temp_resnet,long_term_ar} model spatio-temporal patterns.

As two-stream networks \cite{two_stream} discard the temporal order,  rank pooling \cite{basura_rankpool, basura_rankpool2, anoop_rankpool_nonlin,anoop_advers} and higher-order pooling \cite{hok,me_tensor_eccv16,me_tensor,Pengfei_ICCV19,kon_tpami2020b} are popular.
A recent I3D model \cite{i3d_net} `inflates' 2D CNN filters pre-trained on ImageNet to spatio-temporal 3D filters, and implements temporal pooling. 
PAN~\cite{acmmm19_ZhangZCG19} proposes a motion cue called Persistence of Appearance that enables the network to distill the motion information directly from adjacent RGB frames.  
Approach \cite{acmmm19_LiuGQWL19} uses bootstrapping with long-range temporal context attention while approach \cite{acmmm20_KumarKSXS20} proposes a graph attention model to explore the semantics. 
Slow-I-Fast-P (SIFP) \cite{acmmm20_2020A}  for compressed AR contains the slow and fast pathways I and P, resp., receiving a sparse sampling I-frame clip and  a dense sampling pseudo optical flow clip.

 AssembleNet~\cite{assemblenet}  automatically finds a neural architecture with a good connectivity to capture spatio-temporal interactions for AR through NAS.  
AssembleNet++~\cite{assemblenet_plus} further learns the interactions between raw appearance and/or motion features and spatial object information through learning  dynamic attention weights and search through the inter-block attention connectivity. 

We use DEEP-HAL \cite{Wang_2019_ICCV} which employs a 1D convolution for temporal pooling (I3D net.). We also investigate the use of  AssembleNet and AssembleNet++ as backbones to show that our proposed object and saliency descriptors are independent of the backbone. We focus on the design/ability of ODF/SDF to supervise DEEP-HAL. 

\vspace{0.05cm}
\noindent{\textbf{Power Normalization.}} For BoW/FV and CNN-based streams, the so-called burstiness defined as `{\em the property that a given visual element appears more times in an image than a statistically independent model would predict}' \cite{jegou_bursts} has to be tackled. Thus, we employ Power Normalization~\cite{me_ATN,me_tensor_tech_rep,me_tensor,me_deeper,kon_tpami2020a} which suppresses the burstiness via the so-called MaxExp pooling \cite{me_ATN} given in Section \ref{sec:backgr}.

\section{Background}
\label{sec:backgr}

Below, we present Power Normalization \cite{me_ATN,me_tensor}, count sketches \cite{weinberger_sketch}, and the RBF feature maps  which we use in our pipeline with the goal of the burstiness and dimensionality reduction, and Cartesian coordinate/frame positional encoding. 

\vspace{+0.05cm}
\noindent{\textbf{Notations.}} We use boldface uppercase letters to express matrices \eg, $\mM, \mPP$, regular uppercase letters with a subscript to express matrix elements \eg, $P_{ij}$ is the $(i,j)^{\text{th}}$ element of $\mPP$, boldface lowercase letters to express vectors, \eg $\vx, \vphi, \vpsi$, and regular lowercase letters to denote scalars. Vectors can be numbered \eg, $\vx_n$  while regular lowercase  letters with a subscript express an element of vector \eg, $\vx_i$ is the $i^{\text{th}}$ element of $\vx$. Operators `$;$' and `$,$' concatenate vectors along the first and second mode, respectively,  $\circledcirc_{i\in\idx{K}}\mvv_i\!=[\mvv_1; \cdots; \mvv_K]$ and $\circledcirc^2_{i\in\idx{K}}\mvv_i\!=[\mvv_1, \cdots, \mvv_K]$ concatenate a group of vectors in the first and second mode, respectively,  
$\oplus$ denotes the aggregation (sum) while $\idx{d}$ denotes an index set of integers $\{1,\cdots,d\}$.

\subsection{Power Normalization}
\label{sec:pns}

%
\begin{proposition}
\label{pr:pn}
Sigmoid (SigmE), a Max-pooling approximation \cite{me_deeper}, is an extension of the MaxExp operator defined as $g(\vpsi, \eta)\!=\!1\!-\!(1\!-\!\vpsi)^{\eta}$ for $\eta\!>\!1$ to the operator with a smooth derivative, a response defined for real-valued $\vpsi$ (rather than $\vpsi\!\geq\!0$), a parameter $\eta'$ and a small constant $\epsilon'$:
\vspace{-0.1cm}
\begin{align}
& \!\!\!\!\!g(\vpsi, \eta')\!=\!\frac{2}{1\!+\!\expl{{-\eta'\vpsi}/{(\lVert\vpsi\rVert_2+\epsilon')}}}\!-\!1.
\label{eq:sigmoid}
\end{align}
\begin{proof}
\vspace{-0.15cm}
See papers \cite{me_ATN,me_deeper} for extensive considerations.
\end{proof}
\end{proposition}
As papers \cite{me_ATN,Wang_2019_ICCV} show that various pooling operators perform similarly, we equip our hallucination streams with SigmE followed by count sketching described below.

\subsection{Count Sketches}
\label{sec:sketch}

Sketching vectors by the count sketch \cite{cormode_sketch,weinberger_sketch} is used for their dimensionality reduction which we use in this paper.
\begin{proposition}
\label{pr:ten_sketch}
Let $d$ and $d'$ denote the dimensionality of the input and sketched output vectors, respectively. Let vector $\vh\!\in\!\idx{d'}^d$ contain $d$ uniformly drawn integer numbers from $\{1,\cdots,d'\}$ and vector $\vsss\!\in\!\{-1,1\}^{d}$ contain $d$ uniformly drawn values from $\{-1,1\}$. Then, the sketch projection matrix $\mPP\!\in\!\{-1,0,1\}^{d'\times d}$ becomes:
\vspace{-0.3cm}
\begin{equation}\label{eq:sk1}
P_{ij}\!=\!
\begin{cases} s_i  & \text{if }h_i\!=\!j,
\\
0 &\text{otherwise},
\end{cases}
\end{equation}
and the sketch projection $p: \mbr{d}\!\rightarrow\!\mbr{d'}$  is a linear operation given as $p(\vpsi)\!=\!\mPP\vpsi$ (or $p(\vpsi; \mPP)\!=\!\mPP\vpsi$ to highlight $\mPP$).
\begin{proof}
\vspace{-0.15cm}
It directly follows from the definition of the count sketch \eg, see Definition 1 \cite{weinberger_sketch}.
\end{proof}
\end{proposition}

\begin{remark}
\label{re:ten_sketch}
Count sketches are unbiased estimators:\\ $\mathbb{E}_{\vh,\vsss}(p(\vpsi,\mPP(\vh,\vsss)),p(\vpsi',\mPP(\vh,\vsss)))\!=\!\left<\vpsi,\vpsi'\right>$. As variance\\ $\mathbb{V}_{\vh,\vsss}(p(\vpsi),p(\vpsi'))\!\leq\!\frac{1}{d'}\left(\left<\vpsi,\vpsi'\right>^2 +\lVert\vpsi\rVert_2^2\lVert\vpsi'\rVert_2^2\right)$, the 
 larger sketches are less noisy. Thus, for every modality, we use a separate sketch matrix $\mPP$. 
\begin{proof}
\vspace{-0.15cm}
For the first and second property, see  Appendix A of paper \cite{weinberger_sketch} and Lemma 3 \cite{pham_sketch}.
\end{proof}
\end{remark}

\subsection{Positional Embedding}
\label{sec:kernel_linearization}
Let $G_{\sigma}(\vx\!-\!\vx'\!)=\exp(-\enorm{\vx\!-\!\vx'\!}^2/{2\sigma^2})$ denote a standard Gaussian RBF kernel centered at $\vx'\!$ and having a bandwidth $\sigma$. Kernel linearization refers to rewriting this $G_{\sigma}$ as an inner-product of two infinite-dimensional feature maps. To obtain these maps, we use a fast approximation method based on probability product kernels \cite{jebara_prodkers}. Specifically, we employ the inner product of $d''$-dimensional isotropic Gaussians given $\vx,\vx'\!\!\in\!\mbr{d''}\!$. Thus, we have: 
%
\begin{align}
&\!\!\!\!\!\!\!\!\!\!\!\!G_{\sigma}\!\left(\vx\!-\!\vx'\!\right)\!\!=\!\!\left(\frac{2}{\pi\sigma^2}\right)^{\!\!\frac{d''}{2}}\!\!\!\!\!\!\int\limits_{\vzeta\in\mbr{d''}}\!\!\!\!G_{\sigma/\sqrt{2}}\!\!\left(\vx\!-\!\vzeta\right)G_{\sigma/\sqrt{2}}(\vx'\!\!-\!\vzeta)\,\mathrm{d}\vzeta.
\label{eq:gauss_lin}
\end{align}
Eq. \eqref{eq:gauss_lin} is then approximated by replacing the integral with the sum over $Z$ pivots $\vzeta_1,\cdots,\vzeta_Z$, thus yielding a feature map $\vphi$ as:
\vspace{-0.1cm}
\begin{align}
&\!\!\!\!\!\!\!\!\!\!\!\!\!\!\!\!\!\!\!\!\!\vphi(\vx; \{\vzeta_i\}_{i\in\idx{Z}})=\left[{G}_{\sigma/\sqrt{2}}(\vx-\vzeta_1),\cdots,{G}_{\sigma/\sqrt{2}}(\vx-\vzeta_Z)\right]^T\!\!\!\!,\!\!\!\!\label{eq:gauss_lin2a}\\
\!\!\!\!\!\!\!\!\text{ and } & G_{\sigma}(\vx\!-\!\vx'\!)\approx\left<\sqrt{c}\vphi(\vx), \sqrt{c}\vphi(\vx'\!)\right>,
\label{eq:gauss_lin2}
\end{align}
where $c$ is a const. Eq. \eqref{eq:gauss_lin2} is the linearization of the RBF kernel. Eq. \eqref{eq:gauss_lin2a} is the feature map. $\{\vzeta_i\}_{i\in\idx{Z}}$ are  pivots. As we use 1 dim. signals, we simply cover interval $[0;1]$ with $Z$ equally spaced pivots. For clarity, we drop $\{\vzeta_i\}_{i\in\idx{Z}}$ and write $\vphi(\vx)$, \etc.

\section{Approach}
\label{sec:approach}

Our pipeline is illustrated in Figure \ref{fig:pipe}. It consists of (i) streams already present in DEEP-HAL \cite{Wang_2019_ICCV} such as the FV/BoW streams (black), the High Abstraction Features (HAF) stream and the Optical Flow Features (OFF) which are fed into (ii) the Prediction Network abbreviated as PredNet. In this paper we focus on two non-trivial streams, that is the Object Detection Features and Saliency Detection Features (dashed blue) (ODF and SDF for short).

BoW/FV/OFF streams take the backbone intermediate representations generated from the RGB frames  and learn to hallucinate BoW/FV and the optical flow (I3D only) representations via the MSE loss between the ground-truth BoW/FV/OFF and the outputs of BoW/FV/OFF streams. 
For AssembleNet/AssembleNet++, RGB and optical flow are combined by the backbone, thus we remove the OFF stream. 
The same MSE loss is applied to the ODF and SDF streams. However, the design of compact ground-truth ODF and SDF descriptors is one of our main contributions.

The HAF stream processes the backbone representations prior to combining them with the hallucinated streams. PredNet fuses the combined BoW/FV/OFF/HAF and our new  ODF and SDF to learn actions on videos. 
Below, we start by describing how we obtain our ODF and SDF descriptors before we describe  modules of DEEP-HAL \cite{Wang_2019_ICCV} and our modifications. One change is that we learn weights for the weighted mean pooling (\ie, $\sum_i w'_i\vpsi/\sum_i w'_i$) of each stream to avoid concatenation of streams (prevent overparametrization). 

\subsection{Statistical Motivation}
\label{sec:statmot}
Before we outline our ODF and SDF descriptors, we motivate the use of higher-order statistics. To compare videos, we want to capture a distribution of local features/descriptors \eg, detection scores. The characteristic function $\varphi_\Upsilon(\boldsymbol{\omega})\!=\!\mathbb{E}_{\vupsilon\sim\Upsilon}\left(\exp(\iu\boldsymbol{\omega}^T\!\vupsilon)\right)$ describes the probability density $f_\Upsilon(\vupsilon)$ of some video features (local features $\vupsilon\!\sim\!\Upsilon$). We obtain the  Taylor expansion of the characteristic function:
\vspace{-0.35cm}
\begin{align}
&\!\!\!\!\!\!\!\!\!\mathbb{E}_{\vupsilon\sim\Upsilon}\bigg(\sum\limits_{r=0}^\infty\frac{\iu^j}{r!}\left<\vupsilon,\boldsymbol{\omega}\right>^r\bigg)\!\approx\!\frac{1}{N}\sum\limits_{n=0}^N\sum\limits_{r=0}^\infty\frac{\iu^r}{r!}\left<{\kronstack}_r\vupsilon_n,{\kronstack}_r\boldsymbol{\omega}\right>\!=\!\!\!\!\\
%
&\!\!\!\!\!\sum\limits_{r=0}^\infty\frac{\iu^r}{r!}\bigg<\frac{1}{N}\sum\limits_{n=0}^N{\kronstack}_r\vupsilon_n,{\kronstack}_r\boldsymbol{\omega}\bigg>\!=\!\sum\limits_{r=0}^\infty\left<\tX^{(r)},\frac{\iu^r}{r!}{\kronstack}_r\boldsymbol{\omega}\right>\!,\nonumber
\end{align}

\vspace{-0.35cm}
\noindent
where $\iu$ is the imaginary number, and a tensor descriptor $\tX^{(r)}\!=\!\frac{1}{N}\!\sum\limits_{n=0}^N\!{\kronstack}_r\vupsilon_n$. In principle, with infinite data and infinite moments, one can fully capture  $f_\Upsilon(\vupsilon)$. In practice, first-, second- and third-order moments are typically sufficient, however, second- and third-order tensors grow quadratically and cubically \wrt the size of $\vupsilon$. Thus, in what follows, we represent second-order moments not by a covariance matrix but by the subspace corresponding to the top $n'$ leading eigenvectors. We also make use of the corresponding eigenvalues of the signal. Finally, it suffices to notice that $\vkappa^{(r)}\!=\!\text{diag}\!\left(\tX^{(r)}\right)$ corresponds to the notion of order $r$ cumulants used in calculations of skewness ($r\!=\!3$) and kurtosis ($r\!=\!4$) but it grows linearly \wrt the size of $\vupsilon$. Thus, in what follows, we use the $\ell_2$ norm normalized mean, leading eigenvectors, trace-normalized eigenvalues, skewness and kurtosis (rather than coskewness and cokurtosis) to obtain compact representation of ODF and SDF.

\subsection{Object Detection Features}
\label{sec:odf}

Each object bounding box is described by the  feature vector: 
%
\begin{align}
& \!\!\!\!\!\!\!\!\!\!\!\!\vupsilon\!=\!\left[\vdelta(y_{(det)}); \vy_{(inet)}; \vphi(\varsigma); \circledcirc_{i\in\idx{4}}\vphi(v_i); \vphi\!\left(\frac{\scriptstyle t\!-\!1}{\scriptstyle\tau\!-\!1}\right)\right]\!\in\!\mbr{d}\!\!,
\label{eq:det-feat}
\end{align}
where  $\vdelta\!=[0,\cdots,1,\cdots,0]^T$ is a vector with all zeros but a single $1$ placed at the location $y$. As we have 91 object
classes for detectors trained on the COCO dataset and 80 classes for a detector trained on the AVA
v2.1 dataset, we simply assume $y_{(det)}\!\in\!\idx{91\!+\!80}$, that is, the labels $0,\cdots,91$ describe classes from COCO  while classes $92,\cdots,80\!+\!91$ describe classes from AVA v2.1. Moreover, $\vy_{(inet)}\!\in\!\mbr{1001}$ is an $\ell_1$ norm normalized ImageNet classification score, $0\!\leq\!\varsigma\!\leq\!1$ is the detector confidence score, $v_0,\cdots,v_4$ are the top-left and bottom-right Cartesian coordinates of a bounding box normalized in range $[0;1]$, and $(t\!-\!1)/(\tau\!-\!1)$ is the frame index normalized \wrt the video sequence length $\tau$. For feature maps $\vphi(\cdot)$ defined in Eq. \eqref{eq:gauss_lin2a}, we simply use $Z\!=\!7$ pivots and the  $\sigma$ of RBF is set to $0.5$. Finally, for all detections per video from a given detector, we first compute the mean $\vmu([\vupsilon_1,\cdots,\vupsilon_N])\!\in\!\mbr{d}$ (we write $\vmu$) where $N$ is the total number of detections. Then,  we form a matrix $\mupsilon\!\in\mbr{d\!\times\!N}$:
%
\begin{align}
& \!\!\!\!\!\!\!\!\!\!\!\!\mupsilon\!=\!\frac{\scriptstyle 1}{ \scriptstyle J}\!\left[\frac{\scriptstyle 1}{\scriptstyle K_1}\!\left[\circledcirc^2_{i\in\idx{K_1}}\!(\vupsilon_{i1}\!-\!\vmu)\right],\cdots,\frac{\scriptstyle 1}{\scriptstyle K_J}\!\left[\circledcirc^2_{i\in\idx{K_J}}\!(\vupsilon_{iJ}\!-\!\vmu)\right]\right]\!,\!\!
\label{eq:mat-m}
\end{align}
where $K_j$ denotes a number of detections per frame $j\!\in\!\idx{J}$, 
from which we extract higher-order statistical moments as described below. As $N$ is large and its size varies from video to video, hallucinating $\mupsilon$ directly is not feasible (nor it has invariance 
properties). 

Firstly, we obtain $\mU\mLambda\mV\!=\!\text{svd}\left(\mupsilon\right)$ rather than $\mU\mLambda^2\mU^T\!=\!\text{eig}\left(\mupsilon\!\mupsilon^T\right)$ as $N\!\ll\!d$, where $\mU\!=\![\vu_1,\vu_2,\cdots]$. Take $\tX^{(r)}\!\left(\{\mvv\!\!-\!\!\vmu\}_{n\!=\!0}^N\right)$ (which we abbreviate to $\tX^{(r)}$) and $\vkappa^{(r)}\!=\!\text{diag}\!\left(\tX^{(r)}\right)$ defined in Section \ref{sec:statmot}. We form our multi-moment descriptor $\vpsi_{(det)}\!\in\!\mbr{d(4+n')}$, $n'\!\ge\!1$:
%

\vspace{-0.3cm}
{\fontsize{8.5}{9}\selectfont
\begin{align}
&\label{eq:moment1}\!\!\!\!\!\!\!\!\!\!\vpsi_{(det)}\!=\left[\frac{\vmu}{||\vmu||_2}; \circledcirc^2_{i\in\idx{n'}} \vu_i\left(\tX^{(2)}\right); 
\frac{\vkappa^{(3)}}{\left(\vkappa^{(2)}\right)^{3/2}}; \frac{\vkappa^{(4)}}{\left(\vkappa^{(2)}\right)^{2}};  \frac{\text{diag}(\mLambda^2)}{\sum_i\!\lambda^2_{ii}\!}\!\right]\!,
\end{align}
}
%
%
%

\vspace{-0.1cm}
\noindent
The composition of Eq. \eqref{eq:moment1} is described in Section \ref{sec:statmot}. It is easy to verify that $\frac{\vkappa^{(3)}}{\left(\vkappa^{(2)}\right)^{3/2}}$ and $\frac{\vkappa^{(4)}}{\left(\vkappa^{(2)}\right)^{2}}$ are the empirical versions of skewness and kurtosis  given by $\frac{\mathbb{E}_{\vupsilon\sim\Upsilon}\!\left((\vupsilon\!-\!\vmu)^3\right)}{\mathbb{E}^{3/2}_{\vupsilon\sim\Upsilon}\!\left((\vupsilon\!-\!\vmu)^2\right)}$ and $\frac{\mathbb{E}_{\vupsilon\sim\Upsilon}\!\left((\vupsilon\!-\!\vmu)^4\right)}{\mathbb{E}^{2}_{\vupsilon\sim\Upsilon}\!\left((\vupsilon\!-\!\vmu)^2\right)}$. 

\subsection{Saliency Detection Features}
\label{sec:sdf}

We extract directional gradients from saliency frames by  discretised gradient operators $[-1,0,1]$ and $[-1,0,1]^T$ and obtain  gradient amplitude snd orientation maps $\mLLa$ and $\mTheta$ per frame  encoded by:
%
\begin{align}
& \!\!\!\!\!\!\!\!\!\vupsilon'_{(sal)}=\!\!\!\!\!\!\!\!\!\sum\limits_{i\in\idx{W},j\in\idx{H}}\!\!\!\!\!\!\!\!\Lambda_{ij}\vphi(\theta_{ij}/(2\!\pi))\otimes\vphi\left(\frac{\scriptstyle i\!-\!1}{\scriptstyle  W\!-\!1}\right)\otimes\vphi\left(\frac{\scriptstyle  j\!-\!1}{\scriptstyle  H\!-\!1}\right)\!,
\end{align}
where $\otimes$ is the Kronecker product and $\vphi(\theta)$ follows Eq. \eqref{eq:gauss_lin2a} with the exception that the assignment to Gaussians is realized in the modulo ring to respect the periodical nature of $\theta$. 
We encode $\vphi(\theta)$ with 12 pivots which encode the orientation of gradients. The remaining maps $\vphi(\cdot)$ are encoded with 5 pivots each, which correspond to spatial binning. Note that $\vupsilon'_{(sal)}$ (we write $\vupsilon'$) is similar to a single CKN layer \cite{ckn} but is simpler: for one dimensional variables we sample pivots (\cf learn) for maps $\vphi(\cdot)$. 
Each saliency frame is described as a feature vector $\vupsilon^\dagger\!=\!\left[\scriptstyle\vupsilon'/\scriptstyle||\vupsilon'||_2;\; \scriptstyle\mI_{:}/\scriptstyle||\mI_{:}||_1 \right]\!\in\!\mbr{d^\dagger}\!$, 
%
%
%
where $\mI_{:}$ is a vectorized low-resolution saliency map. Thus, $\vupsilon^\dagger$ captures the directional gradient statistics and the intensity-based gist of saliency maps.
Subsequently, we compute the mean $\vmu([\vupsilon^\dagger_1,\cdots,\vupsilon^\dagger_J])\!\in\!\mbr{d^\dagger}$ (we simply write  $\vmu$) where $J$ is the total number of frames per video. Then, we obtain $\mupsilon^\dagger\!=\!\left[\vupsilon^\dagger_1,\cdots,\vupsilon^\dagger_J\right]\!/\!J\in\!\mbr{d^\dagger\!\times\!J}\!\!$ 
%
%
which is compactly described by the multi-moment  Eq. \eqref{eq:moment1} resulting in $\vpsi_{(sal)}\!\in\!\mbr{d(4+n^\dagger)}$.

\begin{figure}[t]
\centering
\vspace{-0.3cm}
\begin{subfigure}[b]{0.98\linewidth}
\centering\includegraphics[trim=0 0 0 0, clip=true,height=1.8cm]{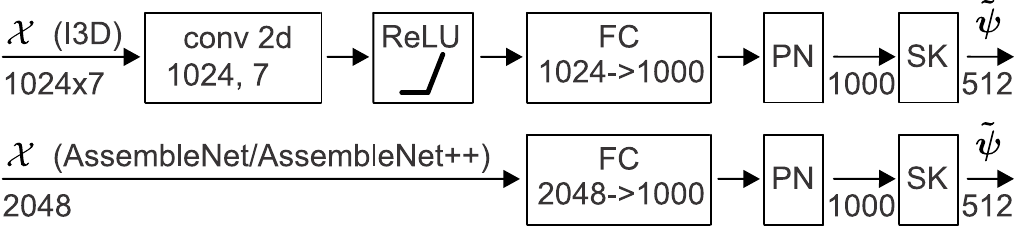}
\vspace{-0.2cm}
\caption{\label{fig:stra}}
\vspace{0.2cm}
\end{subfigure}
\\
\begin{subfigure}[b]{0.98\linewidth}
\centering\includegraphics[trim=0 0 0 0, clip=true,height=0.8cm]{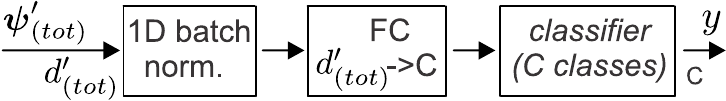}
\vspace{-0.2cm}
\caption{\label{fig:strc}}
\vspace{-0.2cm}
\end{subfigure}
%
\caption{Stream details. Figure \ref{fig:stra} a shows the stream architecture used by us for the FV, BoW, OFF, HAF, DET1$,\cdots,$DET4, SAL1 and SAL2 streams. Figure \ref{fig:strc} shows our PredNet. Operation and their parameters are in each block \eg, {\em conv2d} and its number of filters/size, {\em Power Normalization} ({\em PN}) and {\em Sketching} ({\em SK}). We  indicate the size of input and/or output under arrows.
}\vspace{-0.3cm}
\label{fig:streams}
\end{figure}

\subsection{Hallucinating Streams/High Abstr. Features$\!\!\!\!\!\!\!\!$}
\label{sec:fv-bow}

Each hallucinating stream takes as input the backbone intermediate representation $\mathcal{X}_{(rgb)}$ of size $1024\!\times\!7$ obtained by removing the classifier and the last 1D conv. layer of I3D pre-trained on Kinetics-400. For AssembleNet/AssembleNet++, instead of the classification layer (FC layer), we use a $2048$ dimensional  output from  3D AveragePooling layer.
For the BoW/FV/OFF and HAL streams, we follow the steps described in the DEEP-HAL approach \cite{Wang_2019_ICCV}. For all streams, we use a Fully Connected (FC) unit shown in Figure \ref{fig:stra}. Each stream uses Power Normalization (PN) realized via SigmE and Sketching (SK) from $1000$ to $512$ dim via $\vec{\tilde{\psi}}'_{(\cdot)}\!=\!\tilde{\mPP}_{(\cdot)}\vec{\tilde{\psi}}_{(\cdot)}$. 
Outputs $\vec{\tilde{\psi}}'$ can be now aligned with ground-truth $\vec{\psi}'_{(\cdot)}$ described below. The same steps are applied to High Abstraction Features (HAF), combined with other streams, and also fed into PredNet (see Fig. \ref{fig:pipe}). While hallucinating streams  co-supervise the backbone via external ground-truth tasks, HAF simply passes the backbone features into PredNet.

\vspace{0.05cm}
\noindent{\textbf{Ground-truth BoW/FV/OFF.}} 
We follow the DEEP-HAL setup \cite{Wang_2019_ICCV} and apply PCA to a concatenation of IDT trajectories (30 dim.), HOG (96 dim.), HOF (108 dim.), MBHx (96 dim.) and MBHy (96 dim.). The resulting 213 dim. local descriptors are encoded by FV and BoW with a 256 dim. and a 1000 dim. GMM and k-means dictionaries. For the OFF stream (not used with AssembleNet or AssembleNet++), we pre-computed I3D with LDOF $\mathcal{X}_{(opt.)}$ (Fig. \ref{fig:pipe}). All ground-truth representations were Power Normalized by SigmE/sketched to 512 dim. each via $\vec{\psi}'_{(\cdot)}\!=\!\mPP_{(\cdot)}\vec{\psi}_{(\cdot)}$ and fed into the MSE loss. No ground-truth testing data is used in training/testing. 

\vspace{0.05cm}
\noindent{\textbf{Ground-truth DET1$,\cdots,$DET4/SAL1/SAL2.}} 
The ODF ground-truth training representations are of size $1214\!\times\!N$, where $N$ is the total number of bounding boxes per video (50--10000). The feature dim. 1214 is composed of $80\!+\!91$ dim. one-hot detection classes, $6\!\times\!7$ are the $\vphi(\cdot)$-embedded confidence, bounding box coordinates and the frame number, 1001 is the ImageNet score. We also consider a variant without the RBF embedding: $\vphi(\vx)\!=\!\vx$ ($1178\!\times\!N$ size). The SDF ground-truth training repr. are of size $556\!\times\!J$, where $J$ is the number of frames per video.  300 dim. ($12\!\times\!5\!\times\!5$) concern spatio-angular gradient distributions and 256 dim. ($16\!\times\!16$) concern the luminance of saliency maps. Each ODF/SDF is  encoded per video with the multi-moment descriptor in Eq. \eqref{eq:moment1} yielding  $1178\!\times\!(4+n')$ and  $556\!\times\!(4+n^\dagger)$ compact representations (we vary $n'$ and $n^\dagger$ between 1 and 5). ODF and SDF are Power Normalized by SigmE/sketched to 512 dim. each via $\vec{\psi}'_{(\cdot)}\!=\!\mPP_{(\cdot)}\vec{\psi}_{(\cdot)}$ and fed into the MSE loss. No ground-truth testing representations  were used for training/testing.

\subsection{Objective Function}
\label{sec:obj}

During training, we combine MSE loss functions which co-supervise hallucination streams  with the  classifier:
\vspace{-0.3cm}
\begin{align}
&\;\;\!\!\!\!\!\!\!\!\!\!\!\ell^*(\tX, \vy; \bar{\mP})\!=\!\frac{\alpha}{|\mathcal{H}|}\sum\limits_{i\in\mathcal{H}}{\big\lVert{ \vec{\tilde{\psi}}_i\!-\!\vpsi'_i\big\rVert}}_2^2 \!+\! \ell\!\left(f\!(\vpsi'_{(tot)}; \mP_{(pr)}),\vy; \mP_{(\ell)}\right), \nonumber\\
&\!\!\!\!\!\!\text{ where: } \forall i\!\in\!\mathcal{H}, \vec{\tilde{\psi}}'_i\!=\!\tilde{\mPP}_i g\!\left(\hslash(\tX, \mP_{i}),\eta\right),
\vpsi'_i\!=\!\mPP_i\vpsi_i,\nonumber\\
&\qquad\;\vpsi'_{(haf)}\!=\mPP_{(haf)}g\!\left(\hslash(\tX,\mP_{(haf)}),\eta\right),\nonumber\\
&\qquad\;\vpsi'_{(tot)}\!=\!\frac{1}{|\mathcal{H^*}\!|\!+\!1}\Big(w_{(haf)}\vpsi'_{(haf)}\!\!+\!\!\!\sum_{i\in\mathcal{H^*}}\!w_i\vec{\tilde{\psi}}'_i\Big),\nonumber\\
&\qquad\;\tilde{\vpsi}'_{(det)}\!=\!\frac{1}{|\mathcal{D}|}\!\sum_{i\in\mathcal{D}}\!w_i\vec{\tilde{\psi}}'_i, %
\tilde{\vpsi}'_{(sal)}\!=\!\frac{1}{|\mathcal{S}|}\!\sum_{i\in\mathcal{S}}\!w_i\vec{\tilde{\psi}}'_i.
\label{eq:loss}
\end{align}
The above equation is a trade-off between the MSE loss functions $\{{\lVert{ \vec{\tilde{\psi}}'_i\!-\!\vpsi'_i\rVert}}_2^2, i\!\in\!\mathcal{H}\}$ and the classification loss $\ell(\cdot,\vy; \mP_{(\ell)})$ with some label $\vy\!\in\mathcal{Y}$ and parameters $\mP_{(\ell)}\!\equiv\!\{\mW,\vb\}$. The trade-off is controlled by  $\alpha\!\geq\!0$ while MSE is computed over hall. streams $i\!\in\!\mathcal{H}$, and $\mathcal{H}\!\equiv\!\left\{(fv1),(fv2),(bow),(off),(det1),\cdots,(det4),(sal1),(sal2)\right\}$ is our set of hallucination streams. Moreover, $g(\cdot,\eta)$ is a Power Norm. in Eq. \eqref{eq:sigmoid}, $f(\cdot; \mP_{(pr)})$ is the PredNet module with parameters $\mP_{(pr)}$ which we learn, $\{\hslash(\cdot, \mP_{i}), i\!\in\!\mathcal{H}\}$ are the hallucination streams while $\{\vec{\tilde{\psi}}_i, i\!\in\!\mathcal{H}\}$ are resulting hallucinated BoW/FV/OFF/ODF/SDF representations. We set $\alpha\!=\!1$. Moreover, $\hslash(\cdot, \mP_{(haf)})$ is the HAF stream with the sketched output  $\vpsi'_{(haf)}\!=\!\mPP_{(haf)}\vpsi_{(haf)}$. For the hallucination streams, we learn parameters $\{\mP_{i}, i\!\in\!\mathcal{H}\}$ while for HAF, we learn  $\mP_{{(haf)}}$. The full set of parameters we learn is defined as $\bar{\mP}\!\equiv\!(\{{\mP_{i}, i\!\in\!\mathcal{H}}\}, \mP_{(haf)}, \mP_{(pr)},\mP_{(\ell)})$. Furthermore, $\{\tilde{\mPP}_{i}, i\!\in\!\mathcal{H}\}$ and $\{\mPP_{i}, i\!\in\!\mathcal{H}\}$  are the projection matrices for count sketching of streams $\{\tilde{\vpsi}_{i}, i\!\in\!\mathcal{H}\}$ and the ground-truth feature vectors $\{\vpsi_{i}, i\!\in\!\mathcal{H}\}$. Finally, for $\vpsi'_{(tot)}$ is a weighted average of several streams fed into the PredNet module $f$. 
Moreover, $\mathcal{H^*}\!\equiv\!\left\{(fv1),(fv2),(bow),(off),(det),(sal)\right\}$, $\mathcal{D}\!\equiv\!\left\{(det1),\cdots,(det4)\right\}$ and $\mathcal{S}\!\equiv\!\left\{(sal1),(sal2)\right\}$.
 Section \ref{sec:sketch} details how to select matrices $\mPP$. Let $\mathcal{T}$ be set to either $\mathcal{H^*}$, $\mathcal{D}$ or $\mathcal{S}$, then our weights are:
\vspace{-0.3cm}
\begin{align}
& w_{i}\!=\!\frac{1}{|\mathcal{T}|} \frac{\max(w'^{\beta}_{i}\!\!\!,\rho)}{\sum_{j\in\mathcal{T}}\!\max(w'^{\beta}_j\!\!\!,\rho)}.
\label{eq:wei}
\end{align}
Prior to CNN training, we train an SVM on each ground-truth stream separately (using a manageable training subset), and we set weights $w'$ proportionally to the accuracies obtained on the validation set. For the HAF stream, we simply set $w'_{(haf)}\!=\!\frac{1}{|\mathcal{H^*}\!|\!+\!1}$ and $\rho\!=\!0.1$. For the first few epochs (\ie, 10), we set $\beta\!=\!0$ so that all streams receive equal weights. Subsequently, in each epoch, we run the Golden-section search to find the best $\beta\!\geq\!0$. We start from initial boundary values $\beta\!\in\!\{0,50\}$, we train an SVM on a manageable subset of training data and evaluate $\beta$ on the validation set, 
and we update  boundary values for the next epoch accordingly. Eq. \eqref{eq:wei} has a nice property: for $\beta\!=\!0$, we have $w_i\!=\!1/|\mathcal{T}|$. For $\beta\!\rightarrow\!\infty$, we have $w_i\!=\!1$ if $w_i\!=\!\max(\{w_i\}_{i\in\mathcal{T}})$, otherwise $w_i\!=\!0$. Thus,  $\beta$ interpolates between equalizing all weights and the winner-takes-all solution.
 

\section{Experiments}
\label{sec:exper}

\subsection{Datasets and Evaluation Protocols}
\label{sec:data}

\noindent\textbf{HMDB-51}~\cite{kuehne2011hmdb} has 6766 internet videos/ 51 classes; each video has $\sim$20--1000 frames. We report the mean accuracy across three splits.

\noindent\textbf{YUP++}~\cite{yuppp}  has 20 scene classes of  video textures, 60 videos per class. Splits contain scenes captured by the static or moving camera. We use standard splits (1/9 dataset for training) for evaluation.

\noindent\textbf{MPII Cooking Activities}~\cite{rohrbach2012database} contains high-resolution videos of people cooking dishes. 
The 64  activities from 3748 clips include coarse actions \eg, \emph{opening refrigerator}, and fine-grained actions \eg, \emph{peel}, \emph{slice}, \emph{cut apart}. We use the mean Average Precision (mAP) over 7-fold cross validation. For human-centric protocol \cite{anoop_generalized, anoop_rankpool_nonlin}, we use the faster RCNN \cite{faster-rcnn} to crop video around human subjects.

\noindent\textbf{Charades}~\cite{sigurdsson2016hollywood} consist of of 9848 videos of daily indoors activities, 66500 clip annotations and 157 classes.

\noindent\textbf{EPIC-Kitchens}~\cite{Damen_2018_ECCV} is a multi-class egocentric dataset with ~28K training videos associated with 352 noun and 125 verb classes. The dataset consists of 39,594 segments in 432 videos. 
We follow protocol~\cite{Baradel_2018_ECCV}. 
We evaluate our model on validation, standard seen (S1: 8047 videos), and unseen (S2: 2929 videos) test sets. 

\subsection{Evaluations}
\label{sec:evals}
Below, we show the effectiveness of our method. For smaller datasets, we use the I3D backbone. For  large Charades and EPIC-Kitchens, we additionally investigate  AssembleNet and AssembleNet++ backbones.    
Firstly, we evaluate various design components.

\begin{figure}[b]
\centering
\vspace{-0.1cm}
\begin{subfigure}[b]{0.495\linewidth}
\includegraphics[trim=0 0 0 0, clip=true,width=0.99\linewidth]{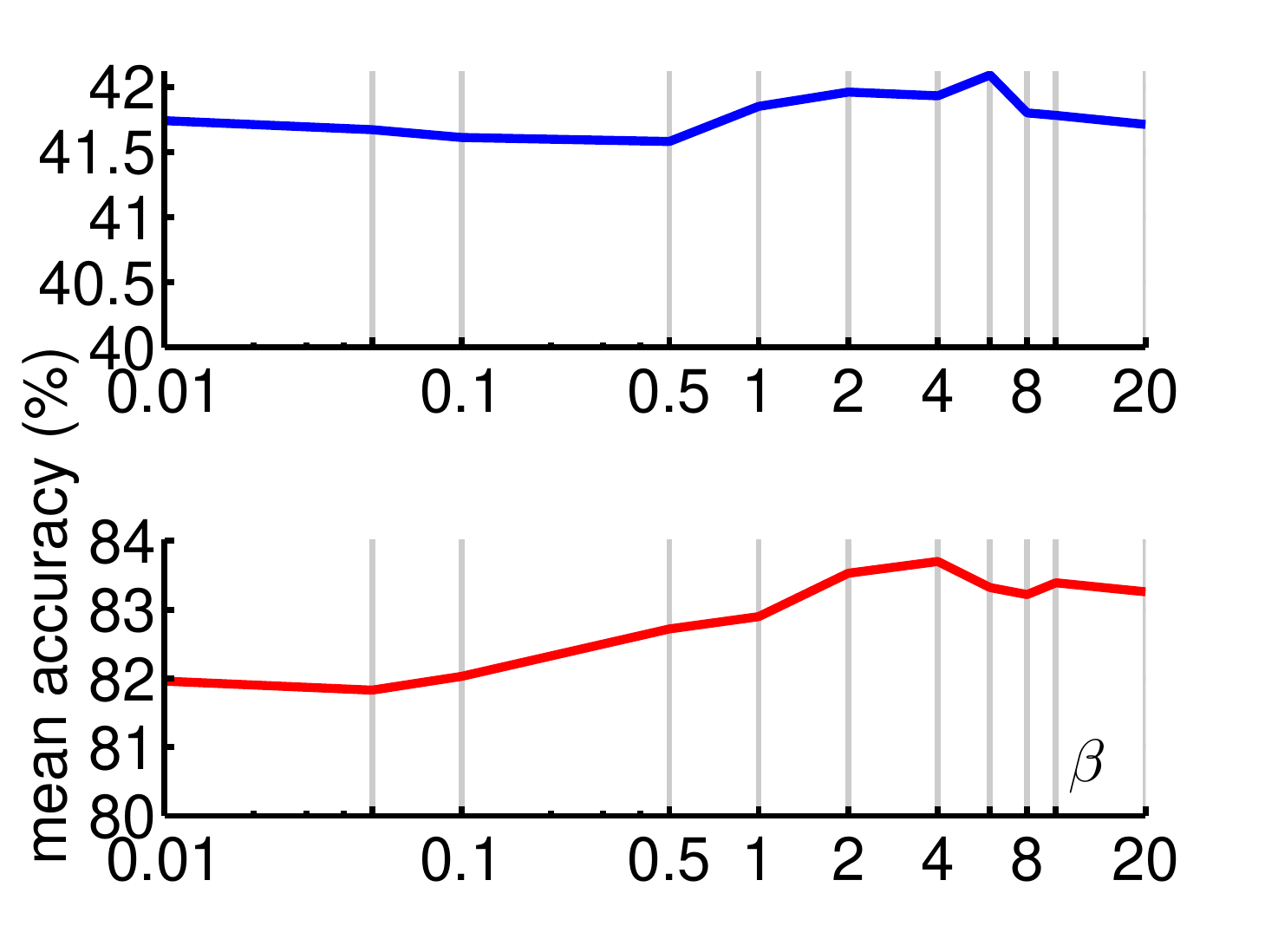}\vspace{-0.2cm}
\caption{\label{fig:beta-h}}
\vspace{-0.2cm}
\end{subfigure}
\begin{subfigure}[b]{0.495\linewidth}
\includegraphics[trim=0 0 0 0, clip=true,width=0.99\linewidth]{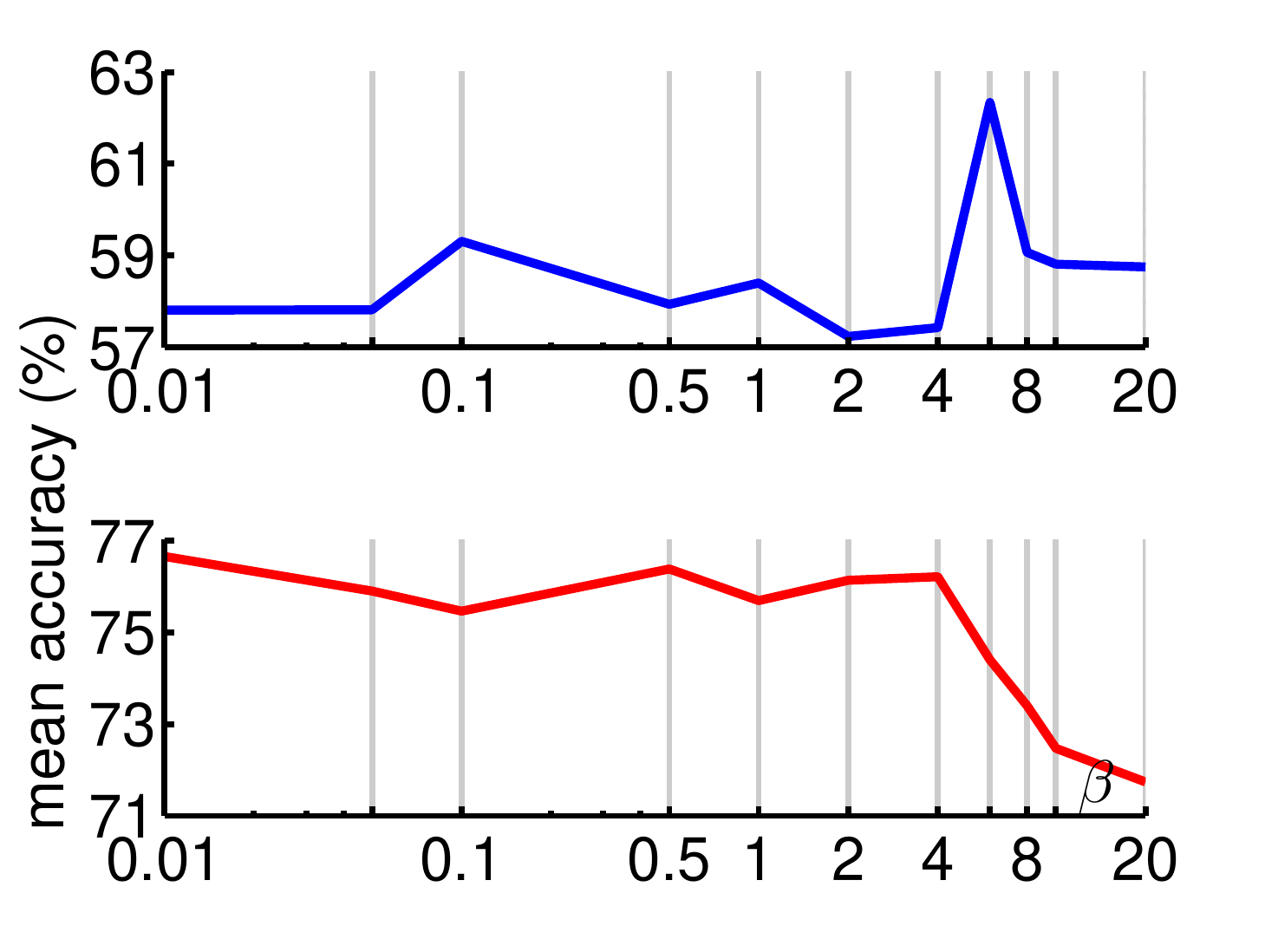}\vspace{-0.2cm}
\caption{\label{fig:beta-y}}
\vspace{-0.2cm}
\end{subfigure}
\caption{The impact of $\beta$ in the weighted mean on the classification results. Figure \ref{fig:beta-h} shows results for HMDB-51 on ({\em top}) four detectors combined+SVM and ({\em bottom}) DEEP-HAL with four detectors combined+SVM. Figure \ref{fig:beta-y} shows results for YUP++.}
\vspace{-0.3cm}
\label{fig:beta}
\end{figure}

\vspace{0.05cm}
\noindent{\textbf{Ground-truth ODF+SVM.}} 
Firstly, we evaluate our ODF on SVM given the HMDB-51 dataset. We set $n'\!=3$ for Eq. \eqref{eq:moment1} and compare various detector backbones and pooling strategies. Table \ref{tab:det1234} shows that all detectors perform similarly with ({\em det3}) being slightly better than other methods. Moreover, max-pooling on ODFs from all four detectors is marginally better than the average-pooling. However, only the weighted mean ({\em all+wei}) according to Eq. \eqref{eq:wei} outperforms ({\em det3}) which highlights the need for the robust aggregation of ODFs. Similarly, when we combine pre-trained DEEP-HAL with all detectors, the weighted mean ({\em DEEP-HAL+all+wei}) performs best. Table \ref{tab:yup-pool} shows the similar trend on YUP++. 
We trained SVM only on videos for which at least one  detection occurred, thus a $75.74\%$ accuracy is much lower than the main results reported on the full pipeline. 
Figure \ref{fig:beta} shows that  $\beta\!\neq\!1$ has a positive impact on reweighting.

\begin{table}[t]
\vspace{-0.3cm}
\parbox{.99\linewidth}{
\setlength{\tabcolsep}{0.12em}
\renewcommand{\arraystretch}{0.70}
\centering
\begin{tabular}{ l c c c c }
\toprule
 & {\em sp1} & {\em sp2} & {\em sp3} & mean acc. \\
\hline
{\em det1} & $42.00\%$ & $39.74\%$ & $40.39\%$ & $40.72\%$\\
{\em det1} & $40.49\%$ & $40.13\%$ & $39.67\%$ & $40.09\%$\\
{\em det3} & $43.78\%$ & $44.05\%$ & $41.97\%$ & $\mathbf{43.26}\%$\\
{\em det4} & $41.08\%$ & $39.22\%$ & $40.39\%$ & $40.23\%$\\
\hdashline
{\em all+avg} & $42.50\%$ & $41.05\%$ & $41.01\%$ & $41.52\%$\\
{\em all+max} & $43.25\%$ & $42.32\%$ & $42.09\%$ & $42.55\%$\\
{\em all+wei} & $45.80\%$ & $44.52\%$ & $44.09\%$ & $\mathbf{44.80}\%$\\
\hdashline
{\em DEEP-HAL+all+avg} & $83.25\%$ & $82.24\%$ & $82.84\%$ & $82.77\%$\\
{\em DEEP-HAL+all+max} & $83.18\%$ & $81.86\%$ & $82.84\%$ & $82.62\%$\\
{\em DEEP-HAL+all+wei} & $84.01\%$ & $83.25\%$ & $83.10\%$ & $\mathbf{83.45}\%$\\
\bottomrule
\end{tabular}
}
\caption{Evaluations of ODF on HMDB-51. ({\em top}) We evaluate backbones such as ({\em det1}) Inception V2, ({\em det2}) Inception ResNet V2, ({\em det3}) ResNet101 and ({\em det4}) NASNet. ({\em middle}) The average-pooled, max-pooled and the weighted mean combination of all detectors are given by ({\em all+avg}), ({\em all+max}) and ({\em all+wei}). ({\em bottom}) Pre-trained DEEP-HAL combined with all four detectors by the average-pooling, max-pooling and the weighted mean.
}
\vspace{-0.3cm}
\label{tab:det1234}
\end{table}
\begin{table}[t]
\vspace{-0.3cm}
\parbox{.99\linewidth}{
\setlength{\tabcolsep}{0.12em}
\renewcommand{\arraystretch}{0.70}
\centering
\begin{tabular}{ l c c c }
\toprule
 & {\em avg} & {\em max} & {\em wei}  \\
\hline
{\em all}          & $55.12\%$ & $42.34\%$ & $\mathbf{60.52}\%$ \\
{\em DEEP-HAL+all} & $74.22\%$ & $71.85\%$ & $\mathbf{75.74}\%$ \\
\bottomrule
\end{tabular}
}
\caption{Pooling on YUP++. Results for the average-pooled ({\em avg}), max-pooled ({\em max}) and the weighted mean  ({\em wei}) of all detectors ({\em all}) \vs pre-trained DEEP-HAL combined with all detectors by the average-pooling, max-pooling and the weighted mean.
}
\vspace{-0.5cm}
\label{tab:yup-pool}
\end{table}

\vspace{0.05cm}
\noindent{\textbf{Ground-truth SDF.}} The SDF on HMDB-51 and YUP++ yielded $24.35\%$ and $32.68\%$ accuracy. This is expected as SDFs do not capture a discriminative information per se but they locate salient spatial and temporal regions to focus the main network on them.

\begin{figure}[b]
\centering
\vspace{-0.1cm}
\begin{subfigure}[b]{0.495\linewidth}
\includegraphics[trim=0 0 0 0, clip=true,width=0.99\linewidth]{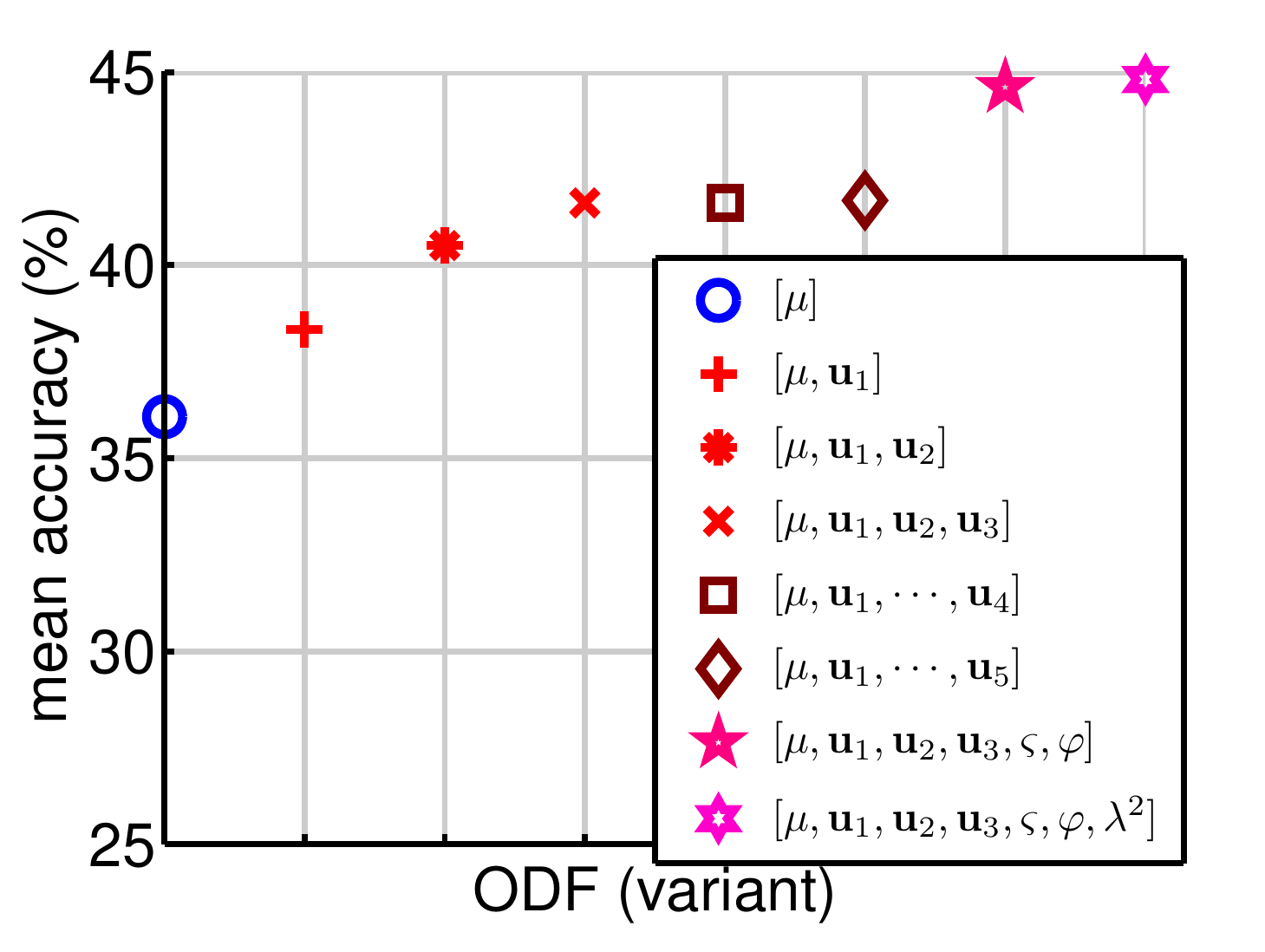}\vspace{-0.2cm}
\caption{\label{fig:odf-h}}
\vspace{-0.2cm}
\end{subfigure}
\begin{subfigure}[b]{0.495\linewidth}
\includegraphics[trim=0 0 0 0, clip=true,width=0.99\linewidth]{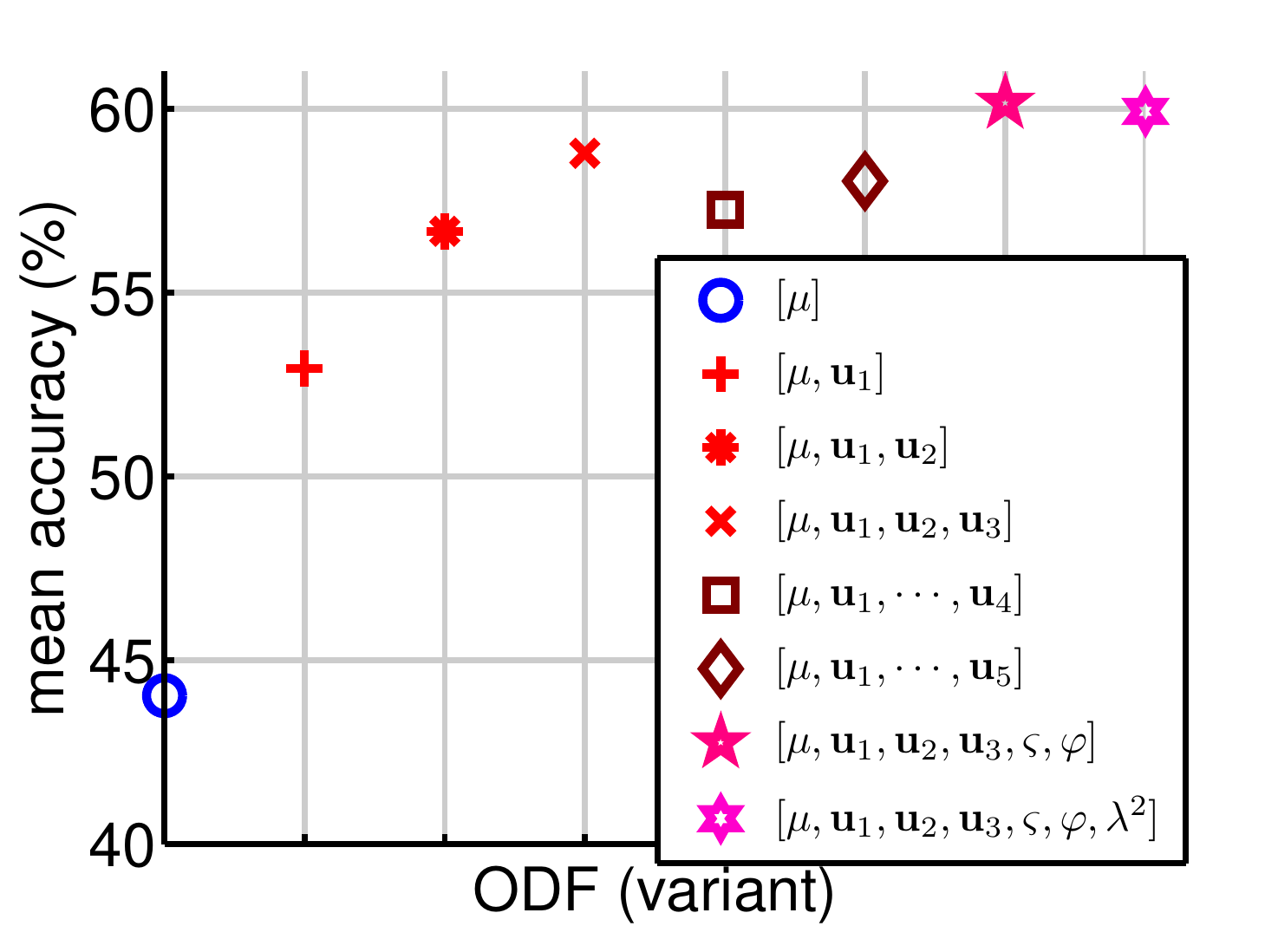}\vspace{-0.2cm}
\caption{\label{fig:odf-y}}
\vspace{-0.2cm}
\end{subfigure}
\caption{ODF eval. on SVM on four detectors (the weighted mean). Fig. \ref{fig:odf-h} and \ref{fig:odf-y}  show results on HMDB-51 and YUP++. $\vmu, \vu_1,\cdots,\vu_i,\boldsymbol{\varsigma}, \boldsymbol{\varphi},$ and $\boldsymbol{\lambda}^2$ correspond to the  entries in Eq. \eqref{eq:moment1}. 
}
\label{fig:odf}
\end{figure}

\vspace{0.05cm}
\noindent{\textbf{Multi-moment descr.}} Figure \ref{fig:odf} shows that the concat. of the mean and three eigenvectors according to Eq. \eqref{eq:moment1} yields good results but adding further vectors deteriorates the performance. Adding skewness and kurtosis ($\boldsymbol{\varsigma}$ and $\boldsymbol{\varphi}$) further improves results, while adding eigenvalues has a limited impact. 

\vspace{0.05cm}
\noindent{\textbf{HMDB-51.}} Table \ref{tab:hmdb51f} shows several  DEEP-HAL variants, which all hallucinate BoW/FV/OFF. DEEP-HAL with  our reweighting mechanism.  ({\em DEEP-HAL+W}) outperforms the original DEEP-HAL denoted as ({\em HAF/BoW/FV hal.}) \cite{Wang_2019_ICCV} by $\sim\!0.8\%$. DEEP-HAL with our ODF and SDF descriptors  ({\em DEEP-HAL+ODF}) and ({\em DEEP-HAL+SDF}) outperform ({\em HAF/BoW/FV hal.}) by $\sim\!1.8\%$ and $\sim\!1.4\%$, resp. This shows that both ODF and SDF are effective. 
Combining DEEP-HAL, ODF and SDF  outperform DEEP-HAL by $\sim\!2.7\%$ demonstrating the complementary nature of ODF and SDF. Utilizing our weighting mechanism with  DEEP-HAL, ODF and SDF denoted as ({\em DEEP-HAL+W+ODF+SDF})  outperform ({\em HAF/BoW/FV hal.}) by $\sim\!4.6\%$. Finally, DEEP-HAL with weighting, and ODF and SDF with RBF feature maps from Eq. \eqref{eq:gauss_lin2a} outperform ({\em HAF/BoW/FV hal.}) by $\sim\!5.1\%$.

\vspace{0.05cm}
\noindent{\textbf{YUP++.}} Table \ref{tab:yupf} shows that ODF is better than SDF, that is ({\em DEEP-HAL+ODF}) and ({\em DEEP-HAL+SDF}) outperform ({\em HAF/BoW/FV hal.}) by $\sim\!0.6\%$ and $\sim\!0.2\%$, resp. This is expected as YUP++ contains dynamic scenes without objects/specific saliency regions correlating with class concepts. However, a combination of detectors/saliency ({\em DEEP-HAL+SDF}) plus weighting ({\em DEEP-HAL+W+ODF+SDF}) plus the RBF  maps ({\em DEEP-HAL+W+G+ODF+SDF}) outperform ({\em HAF/BoW/FV hal.}) by $\sim\!0.7\%$, $\sim\!1.6\%$ and $\sim\!1.8\%$ accuracy, resp.

\vspace{0.05cm}
\noindent{\textbf{MPII.}} Table \ref{tab:mpiif} shows a $\sim\!3.0\%$ mAP gain over ({\em HAF/BoW/FV hal.}) due to detectors capturing the human interaction with objects. 

\begin{table}[t]
\parbox{.99\linewidth}{
\setlength{\tabcolsep}{0.12em}
\renewcommand{\arraystretch}{0.70}
\centering
\begin{tabular}{ l c c c c }
\toprule
 & {\em sp1} & {\em sp2} & {\em sp3} & mean acc. \\
\hline
{\em DEEP-HAL+W} & $83.94\%$ & $82.50\%$ & $83.34\%$ & $83.26\%$\\
{\em DEEP-HAL+ODF} & $85.03\%$ & $83.59\%$ & $84.25\%$ & $84.29\%$\\
{\em DEEP-HAL+SDF} & $84.64\%$ & $83.20\%$ & $83.82\%$ & $83.88\%$\\
{\em DEEP-HAL+ODF+SDF} & $86.14\%$ & $83.66\%$ & $85.81\%$ & $85.20\%$\\
{\em DEEP-HAL+W+ODF+SDF} & $87.78\%$ & $86.27\%$ & $87.06\%$ & $87.04\%$\\
{\em DEEP-HAL+W+G+ODF+SDF}  & $88.37\%$ & $86.80\%$ & $87.52\%$ & $\mathbf{87.56}\%$\\
\midrule
\end{tabular}
}
\parbox{.99\linewidth}{
\setlength{\tabcolsep}{0.12em}
\renewcommand{\arraystretch}{0.70}
\fontsize{9}{9}\selectfont
\centering
\begin{tabular}{ c c }
\kern-0.5em ADL+I3D $81.5\%$ \cite{anoop_advers} &  Full-FT I3D $81.3\%$ \cite{i3d_net}\\
\kern-0.5em EvaNet (Ensemble) $82.3\%$ \cite{Piergiovanni_2019_ICCV} & PA3D + I3D $82.1\%$ \cite{Yan_2019_CVPR}\\
 HAF/BoW/FV exact $82.50\%$ \cite{Wang_2019_ICCV} & HAF/BoW/FV hal. $82.48\%$ \cite{Wang_2019_ICCV}\\
\bottomrule
\end{tabular}
}
\caption{Evaluations of ({\em top}) our methods and ({\em bottom}) comparisons to the state of the art on  HMDB-51.}
\vspace{-0.3cm}
\label{tab:hmdb51f}
\end{table}

\begin{table}[t]
\hspace{-0.45cm}
\parbox{.99\linewidth}{
\setlength{\tabcolsep}{0.12em}
\renewcommand{\arraystretch}{0.70}
\centering
\begin{tabular}{ l c c c c c }
\toprule
 & \multirow{2}{*}{\em static} & \multirow{2}{*}{\em dynamic} & \multirow{2}{*}{\em mixed} & mean & mean \\
&                &             &             & {\fontsize{8}{9}\selectfont stat/dyn}  & all \\
\hline
{\em DEEP-HAL+ODF}    				& $95.00\%$ & $90.93\%$ & $93.52\%$ & $93.0\%$ & $93.2\%$ \\
{\em DEEP-HAL+SDF}    				& $94.96\%$ & $89.93\%$ & $93.58\%$ & $92.4\%$ & $92.8\%$ \\
 {\fontsize{8}{9}\selectfont {\em DEEP-HAL+SDF+ODF}}  & $95.10\%$ & $91.11\%$ & $93.61\%$ & $93.1\%$ & $93.3\%$ \\
 {\fontsize{8}{9}\selectfont {\em DEEP-HAL+W+SDF+ODF}}  & $96.30\%$ & $92.22\%$ & $94.17\%$ & $94.3\%$ & $94.2\%$ \\
 {\fontsize{8}{9}\selectfont {\em DEEP-HAL+W+G+SDF+ODF}}  & $96.30\%$ & $\mathbf{92.40}\%$ & $\mathbf{94.35}\%$ & $\mathbf{94.4}\%$ & $\mathbf{94.4}\%$ \\
\midrule
T-ResNet \cite{yuppp} & $92.4\%$ & $81.5$\% & $89.0$\% & $87.0\%$ & $87.6\%$\\
ADL I3D \cite{anoop_advers} & $95.1\%$ & $88.3$\% & - & $91.7\%$ & -\\
HAF/BoW/FV hal. \cite{Wang_2019_ICCV} & $94.8\%$ & $89.6$\% & $93.3$\% & $92.2\%$ & $92.6$\%\\
MSOE-two-stream \cite{Hadji_2018_ECCV} & $\mathbf{ 97.0}\%$ & $87.0$\% & $91.8$\% & $92.0\%$ & $91.9$\%\\
\bottomrule
\end{tabular}
}
\caption{Evaluations of ({\em top}) our methods and ({\em bottom}) comparisons to the state of the art on YUP++.}
\vspace{-0.3cm}
\label{tab:yupf}
\end{table}

\begin{table}[!tb]
\parbox{.99\linewidth}{
\setlength{\tabcolsep}{0.12em}
\renewcommand{\arraystretch}{0.70}
\centering
\begin{tabular}{ l c c c c c c c c }
\toprule
 & {\em sp1} & {\em sp2} & {\em sp3} & {\em sp4} & {\em sp5} & {\em sp6} & {\em sp7} & mAP \\
\hline
{\fontsize{7.5}{9}\selectfont {\em DEEP-HAL+W+ODF+SDF}}      		 & $82.5$ & $85.1$ & $85.6$ & $83.5$ & $86.6$ & $80.8$ & $81.2$ & $83.6\%$\\
{\fontsize{7.5}{9}\selectfont {\em DEEP-HAL+W+G+ODF+SDF}}     & $83.3$ & $87.6$ & $85.6$ & $83.4$ & $86.6$ & $83.2$ & $83.6$ & $\mathbf{84.8}\%$\\
\midrule
\end{tabular}
}
\parbox{.99\linewidth}{
\centering
\setlength{\tabcolsep}{0.12em}
\renewcommand{\arraystretch}{0.70}
\fontsize{8}{9}\selectfont
\begin{tabular}{ c c }
\kern-0.5em KRP-FS+IDT $76.1\%$ \cite{anoop_rankpool_nonlin} & GRP+IDT $75.5\%$ \cite{anoop_generalized}\kern-0.5em\\
\kern-0.5em I3D+BoW/OFF MTL $79.1\%$ \cite{Wang_2019_ICCV} & HAF/BoW/OFF hal. $81.8\%$ \cite{Wang_2019_ICCV}\\
\bottomrule
\end{tabular}
}
\caption{Evaluations of ({\em top}) our methods and ({\em bottom}) comparisons to the state of the art on MPII.}
\vspace{-0.3cm}
\label{tab:mpiif}
\end{table}

\vspace{0.05cm}
\noindent{\textbf{Charades.}} Table \ref{tab:charades} (top) presents relative gains of our hallucination pipeline ({\em DEEP-HAL}) with weighted mean pooling ({\em W}) and the RBF maps ({\em G}) denoted as ({\em DEEP-HAL+W+G}). We evaluate Object Detection Features ({\em ODF}) and Saliency Detection Features ({\em SDF}) with 512 dim. sketching ({\em SK512})  
and note that ({\em DEEP-HAL+W+G+ODF (SK512)}) outperforms ({\em DEEP-HAL+W+G+SDF (SK512)}), and both methods outperform the baseline ({\em HAF/BoW/FV hal.}) \cite{Wang_2019_ICCV}. 

Table \ref{tab:charades} (bottom) shows that combining ODF and SDF into ({\em DEEP-HAL+W+G+SDF+ODF (SK512)}) yields $49.06\%$ mAP which constitutes on a $\sim\!\mathbf{6}\%$ gain over the baseline({\em HAF/BoW/FV hal.}) \cite{Wang_2019_ICCV}. This demonstrates that ODF and SDF are highly complementary. Applying a larger sketch ({\em DEEP-HAL+W+G+ODF+SDF (SK1024)}) yields $50.14\%$ mAP which matches the use  ({\em DEEP-HAL+W+G+ODF+SDF (exact)}) that denotes a late fusion by concatenation of ODF and SDF with the stream resulting from DEEP-HAL fed into PredNet. Note that ({\em exact}) indicates that ODF and SDF are not hallucinated at the test time but they are computed. 
the results matching between ({\em DEEP-HAL+W+G+ODF+SDF (SK1024)}) and  ({\em DEEP-HAL+W+G+ODF+SDF (exact)}) show that we can hallucinate ODF and SDF at the test time while regaining the full performance. We save computational time and hallucinate the detection and saliency features which boost results on Charades by $\sim\!\mathbf{6}\%$ over the baseline. 

Table \ref{tab:charades2} shows that our idea applied to AssembleNet and AssembleNet++ yields state of the art \eg, we outperform these two networks by \textbf{4.5}\% and \textbf{5.6}\% mAP, respectively. We note that our detectors do not need to be computed at all at the test time. 

In contrast, the best currently reported papers such as SlowFast networks \cite{slowfast} and AssembleNet \cite{assemblenet} achieve 45.2\% and 51.6\% on Charades. 
As SlowFast networks and AssembleNet backbones can be used in place of I3D in our experimental setup,  our approach is `orthogonal' to these latest developments  which focus on heavy mining for combinations of neural blocks/dataflow between them to obtain an `optimal' pipeline. 
We achieve similar results with a simple approach based on self-supervised learning. Our pipeline is lightweight by comparison (no need for computations of the optical flow, or detections or segmentation masks at test time).

\noindent{\textbf{ImageNet (global score) \vs object detectors.}} 
Various scores from the object and saliency detectors which we use cannot be plugged directly into the DEEP-HAL due to the varying number of objects detected and the varying number of frames, thus we propose and use ODF and SDF descriptors. We also note that using a simplified variant of ODF which stacks up ImageNet scores per frame into a matrix (no detectors) to which we apply our multi-moment descriptor yielded $\sim\!4\%$ worse results on Charades than our DEEP-HAL+ODF (detectors-based approach) which yields $48.0\%$ mAP. This is expected as ImageNet is trained in a multi-class setting (one object per image) while detectors let us model robustly distributions of object classes and locations per frame.

\vspace{0.05cm}
\noindent{\textbf{EPIC-Kitchens.}}Table~\ref{epic-kitchens} shows the experimental results. I3D and AssembleNet/AssembleNet++ learn human-like semantic features due to ODF/SDF, and there is no evidence a backbone can discover these without a guidance. By comparing MPII (3748 clips) with large EPIC-Kitchens (39594 clips) (both about cooking), SDF+ODF boost MPII from 81.8 to 84.8\%, and SDF+ODF boost EPIC-Kitchens from 32.51\% (DEEP-HAL) to 35.88\% (on seen classes protocol), and from 22.33\% (DEEP-HAL) to 27.32\% (on unseen classes protocol). The boost is ~\textbf{3}\% on both MPII and EPIC-Kitchens (nearly 10$\times$ more clips than MPII).

\newcommand{\fsnine}[0]{\fontsize{9}{9}\selectfont}
\newcommand{\fsninee}[0]{\fontsize{9}{9}\selectfont}
\begin{table}[t]
\parbox{.99\linewidth}{
\setlength{\tabcolsep}{0.12em}
\renewcommand{\arraystretch}{0.70}
\centering
\begin{tabular}{ c c c }
\toprule
\fsninee HAF/BoW/FV     				& \fsninee {\em DEEP-HAL+}         & \fsninee {\em DEEP-HAL+}     \\
\fsninee hal. \cite{Wang_2019_ICCV}			&  \fsninee {\em W+G+ODF (SK512)}         & \fsninee {\em W+G+SDF (SK512)}  \\
\hline
\fsnine 43.1 & \fsnine 47.22 & \fsnine 45.30   \\
\midrule
\end{tabular}
}
\parbox{.99\linewidth}{
\setlength{\tabcolsep}{0.12em}
\renewcommand{\arraystretch}{0.70}
\centering
\begin{tabular}{ c c c }
\fsninee {\em DEEP-HAL+W+G+}      						& \fsninee {\em DEEP-HAL+W+G+}         				   & \fsninee {\em DEEP-HAL+W+G+}   \\
\fsninee {\em ODF+SDF (SK512)}			& \fsninee {\em ODF+SDF (SK1024)}       & \fsninee {\em ODF+SDF (exact)} \\
\hline
\fsnine 49.06 & \fsnine {\bf 50.14} & \fsnine {\bf 50.16} \\
\bottomrule
\end{tabular}
}
\caption{Evaluations of our methods on  Charades (I3D backbone).}
\label{tab:charades}
\end{table}

\begin{table}[t]
\parbox{.99\linewidth}{
\setlength{\tabcolsep}{0.12em}
\renewcommand{\arraystretch}{0.70}
\centering
\begin{tabular}{c c c c }
\toprule
\multicolumn{4}{c}{{\em AssembleNet++ 50} (Kinetics-400 pre-training)}\\
\fsninee baseline &
\fsninee {\em ODF+SDF (SK512)}			& \fsninee {\em ODF+SDF (SK1024)}       & \fsninee {\em ODF+SDF (exact)} \\
\hline
\fsnine 53.8 & \fsnine 55.81 & \fsnine {\bf 56.94} & \fsnine {\bf 57.30} \\
\midrule
\end{tabular}
}

\parbox{.99\linewidth}{
\setlength{\tabcolsep}{0.12em}
\renewcommand{\arraystretch}{0.70}
\centering
\begin{tabular}{ c c c c }


\multicolumn{4}{c}{{\em AssembleNet++ 50} (without pre-training)}\\
\fsninee baseline &
\fsninee {\em ODF+SDF (SK512)}			& \fsninee {\em ODF+SDF (SK1024)}       & \fsninee {\em ODF+SDF (exact)} \\

\hline
\fsnine 56.7 & \fsnine 60.71 & \fsnine \textbf{61.98} & \fsnine \textbf{62.29} \\
\bottomrule
\end{tabular}
}
\caption{Evaluations of our methods on the Charades dataset (AssembleNet and AssembleNet++ backbones). Note that we do not use segmentation masks for AssembleNet and AssembleNet++, thus baseline results reported by us are slightly lower compared to authors' results of 55.0\% and 59.8\% mAP, respectively.}
\label{tab:charades2}
\end{table}

\begin{table}[!ht]
\begin{center}
\resizebox{0.80\textwidth}{!}{\begin{tabular}{ l c  c c  c c  c }
\toprule
& \multicolumn{2}{c}{Verbs} & \multicolumn{2}{c}{Nouns} & \multicolumn{2}{c}{Actions}\\
\cline{1-7}
& top-1 & top-5 & top-1 & top-5 & top-1 & top-5\\
\hline
& \multicolumn{6}{c}{\bf Validation}\\
LFB Max~\cite{Wu_2019_CVPR} & 52.6 & 81.2 & 31.8 & 56.8 & 22.8 & 41.1\\
WeakLargeScale~\cite{Ghadiyaram_2019_CVPR} & 58.4 & 84.1 & 36.9 & 60.3 & 26.1 & 42.7\\
\hdashline
DEEP-HAL+ODF+SDF(SK1024) & 55.4 & 82.9 & 33.3 & 55.1 & 21.5 & 39.7\\
AssembleNet++ ODF+SDF(SK512) & 57.2 & 84.6 & 34.8 & 56.4 & 23.2 & 41.3\\
AssembleNet++ ODF+SDF(SK1024) & 58.7 & 85.6 & 36.0 & 57.3 & {\bf 24.7} & {\bf 43.0} \\
\hdashline
AssembleNet++ ODF+SDF(exact) & 60.0 & 86.7 & 37.1 & 59.2 & {\bf 25.2} & {\bf 43.4} \\

\midrule
& \multicolumn{6}{c}{\bf Test s1 (seen)}\\
TSN Fusion~\cite{Damen_2018_ECCV} & 48.2 & 84.1 & 36.7 & 62.3 & 20.5 & 39.8\\
LFB Max~\cite{Wu_2019_CVPR} & 60.0 & 88.4 & 45.0 & 71.8 & 32.7 & 55.3\\
WeakLargeScale~\cite{Ghadiyaram_2019_CVPR} & 65.2 & 87.4 & 45.1 & 67.8 & 34.5 & 53.8\\
\hdashline
DEEP-HAL+ODF+SDF(SK1024) & 62.2 & 85.0 & 46.1 & 69.3 & 32.5 & 53.6\\
AssembleNet++ ODF+SDF(SK1024) & 65.0 & 87.8 & 48.8 & 72.5 & {\bf 35.0} & {\bf 56.1}\\
\hdashline
AssembleNet++ ODF+SDF(exact) & 66.2 & 88.5 & 49.3 & 72.8 & {\bf 35.8} & {\bf 56.8}\\
\midrule
& \multicolumn{6}{c}{\bf Test s2 (unseen)}\\

TSN Fusion~\cite{Damen_2018_ECCV} & 39.4 & 74.3 & 22.7 & 45.7 & 10.9 & 25.3\\
LFB Max~\cite{Wu_2019_CVPR} & 50.9 & 77.6 & 31.5 & 57.8 & 21.2 & 39.4\\
WeakLargeScale~\cite{Ghadiyaram_2019_CVPR} & 57.3 & 81.1 & 35.7 & 58.7 & 25.6 & 42.7\\
\hdashline
DEEP-HAL+ODF+SDF(SK1024) & 55.3 & 79.1 & 32.6 & 55.4 & 22.3 & 39.2\\
AssembleNet++ ODF+SDF(SK1024) & 58.3 & 82.1 & 35.2 & 58.2 & {\bf 25.9} & {\bf 42.9}\\
\hdashline
AssembleNet++ ODF+SDF(exact) & 59.0 & 83.3 & 35.7 & 59.0 & {\bf 27.3} & {\bf 44.0}\\

\bottomrule
\end{tabular}}
\end{center}
\caption{Experimental results on the EPIC-Kitchens.
}
\label{epic-kitchens}
\vspace{-0.3cm}
\end{table}

\section{Conclusions}
\label{sec:concl}

We have introduced two simple yet effective object and saliency descriptors, which perform self-supervision of an AR hallucination-based network. We have shown that modeling high-order statistical moments can result in small representations that can self-supervise our AR pipeline. The findings are in line with recent multi-task learning papers which argue that related tasks can co-supervise the main task. We are the first to hallucinate object and saliency detection descriptors with clear cut improvements in accuracy, and state-of-the-art results on the large-scale Charades and EPIC-Kitchens. More importantly, we demonstrate that hallucinating object and saliency detections is an attractive proposition even for the state-of-the-art AR backbones  such as AssembleNet and AssembleNet++.

\appendix 
\addcontentsline{toc}{chapter}{APPENDICES}
\begin{table}[t]
\vspace{0.3cm}
\parbox{.99\linewidth}{
\setlength{\tabcolsep}{0.12em}
\renewcommand{\arraystretch}{0.70}
\centering
\begin{tabular}{ l c c c c }
\toprule
 & {\em sp1} & {\em sp2} & {\em sp3} & mean acc. \\
\hline
{\em wei+flat} & $86.47\%$ & $85.56\%$ & $86.27\%$ & $86.10\%$\\
{\em wei+3 levels} & $88.37\%$ & $86.80\%$ & $87.52\%$ & $\mathbf{87.56}\%$\\
\bottomrule
\end{tabular}
}
\caption{Evaluations of the flat single level weighted mean ({\em wei+flat}) \vs three levels of weighted mean pooling ({\em wei+3 levels}) on HMDB-51.
}
\label{tab:weilevel}
\end{table}

\section{Reweighting mechanism}

In this experiment, we employ pipeline ({\em DEEP-HAL+W+G+SDF+ODF (SK512)}) explained above. Typically, we use three levels of weighting mean pooling which are applied to (i) four object detectors constituting on ODF, (ii) two saliency detectors constituting on SDF, and (iii) the final combination of HAF/BOW/FV/OFF/ODF/SDF. Thus, below we investigate the performance of a single weighting mean pooling  step applied simultaneously to four object detectors, two saliency detectors and the remaining streams. 

Table \ref{tab:weilevel} shows that using a flat single level weighted mean pooling yields 86.1\% accuracy on the HMDB-51 which is a $\sim\!1.4\%$ less compared to utilizing  three levels of weighted mean pooling. 
We expect that having one weighted mean pooling per modality is a reasonable strategy as for instance object category detectors may yield similar responses thus they should be first reweighted for the best `combined detector' performance before being combined with highly complementary modalities.

Finally, Figure \ref{fig:gold} (top) demonstrates how our Golden-search selects optimal $\beta$ on the validation set of MPII ({\em split1}).  Figure \ref{fig:gold} (bottom) demonstrates the corresponding validation mAP (this is not the mAP score on the testing set). Note that for the first 10 epochs we use $\beta\!=\!0$ and we start the Golden-search from epoch 11.

\section{Dataset statistics and timing}

\begin{table}[t]
\vspace{-0.3cm}
\parbox{.99\linewidth}{
\setlength{\tabcolsep}{0.12em}
\renewcommand{\arraystretch}{0.70}
\centering
\begin{tabular}{ l c c c c c }
\toprule
 & {\em no. of} & {\em av. frame} & {\em no. of} & {\em no. of} & {\em no. of} \\
 & {\em frames} & {\em count} & {\em  videos} & {\em  clips} & {\em classes} \\
\hline
HMDB-51  & 628635		& 92.91  & 6766 & 6766 & 51\\
YUP++		 & 166463		& 138.72 & 1200 & 1200 & 20 \\
MPII		 & 662394		& 176.73 & 44   & 3748 & 60 \\
Charades & 19978821 & 300.51 & 9848 & 66500 & 157\\
\bottomrule
\end{tabular}
}
\caption{Statistics of datasets used in our experiemnts.
}
\vspace{-0.3cm}
\label{tab:datasets}
\end{table}

Table \ref{tab:datasets} shows basic statistics re. datasets  used in our experiments. We note that Charades with 66500 uniquely annotated clips, 157 action labels and an average frame count of $300$ per clip is the largest among these datasets.

\begin{table}[t]
\parbox{.99\linewidth}{
\setlength{\tabcolsep}{0.12em}
\renewcommand{\arraystretch}{0.70}
\centering
\begin{tabular}{ l c c c c c }
\toprule
 & \fsninee {\em DET1:} & \fsninee {\em DET2:} & \fsninee {\em DET3:} & \fsninee {\em DET4:} & \fsninee ODF  \\
 & \fsninee {\em Inception} & \fsninee {\em Inception} & \fsninee {\em ResNet101} & \fsninee {\em NASNet} & \fsninee total  \\
 & \fsninee {\em V2} & \fsninee {\em ResNet V2} & \fsninee {\em  AVA} &  & \fsninee (+SVD)  \\
\hline
{\em sec. per frame}   & 0.07		& 0.38  & 0.10 & 0.91 & 1.46 (+0.09) \\
\hline
{\em s.p.c.} HMDB-51   & 6.5		& 35.3  & 9.3 & 84.5 		& 135.6 (+0.5)\\
{\em s.p.c.} YUP++     & 9.7		& 52.7  & 13.9 & 126.2  & 202.5 (+0.8)\\
{\em s.p.c.} MPII      & 12.4		& 67.1  & 17.7 & 160.8  & 258.0 (+1.3)\\
{\em s.p.c.} Charades  & 21.0		& 114.2  & 30.0 & 273.5 & 438.7 (+2.6)\\
\bottomrule
\end{tabular}
}
\caption{Statistics of object detectors we use. We provide timings such as  seconds per frame ({\em sec. per frame}) and seconds per clip ({\em s.p.c.}) for detectors used by ODF. The total time incurred by a combined detector ({\em ODF total}) is also provided. We also compute the time taken by the full SVD and all remaining ODF operations described in the main paper, assuming $\sim\!5$ detections per frame.
}
\label{tab:stat1}
\end{table}
%

%
\begin{table}[t]
\hspace{-0.3cm}
\parbox{.99\linewidth}{
\setlength{\tabcolsep}{0.12em}
\renewcommand{\arraystretch}{0.70}
\centering
\begin{tabular}{ l c c c c }
\toprule
 & \fsninee {\em SAL1:} & \fsninee {\em SAL2:} & \fsninee SDF           & \fsninee ODF+SDF  \\
 & \fsninee {\em MNL} & \fsninee {\em ACLNet}  & \fsninee {\em total}   & \fsninee total  \\
 &                   &                         & \fsninee (+Eq. (10))  &\fsninee (+Eq. (10)+SVD)  \\
\hline
{\em sec. per frame}   & 0.60		& 0.30  & 0.90 (+0.003) & 2.36 (+0.1)  \\
\hline
{\em s.p.c.} HMDB-51   & 55.7		& 27.9  &  83.6 (+0.3) & 219.2 (+0.8) \\
{\em s.p.c.} YUP++     & 83.2		& 41.6  & 124.8 (+0.4) & 327.3 (+1.2) \\
{\em s.p.c.} MPII      & 106.0	& 53.0  & 159.0 (+0.5) & 417.0 (+1.8) \\
{\em s.p.c.} Charades  & 180.3	& 90.1  & 270.4 (+0.9) & 709.1 (+3.5) \\
\bottomrule
\end{tabular}
}
\caption{Statistics of saliency detectors we use. We provide timings such as  seconds per frame ({\em sec. per frame}) and seconds per clip ({\em s.p.c.}) for detectors used by SDF. The total time incurred by a combined detector ({\em SDF total}) is also provided. We also compute the time taken by the descriptor in Eq. (10) and all remaining SDF operations described in the main paper. Finally, we also provide the combined ODF and SDF time ({\em SDF+ODF total}).
}
\vspace{-0.3cm}
\label{tab:stat2}
\end{table}

\begin{figure*}[!th]
\centering
\vspace{-0.3cm}
\begin{subfigure}[b]{0.48\linewidth}
\includegraphics[trim=0 0 0 0, clip=true,height=6cm]{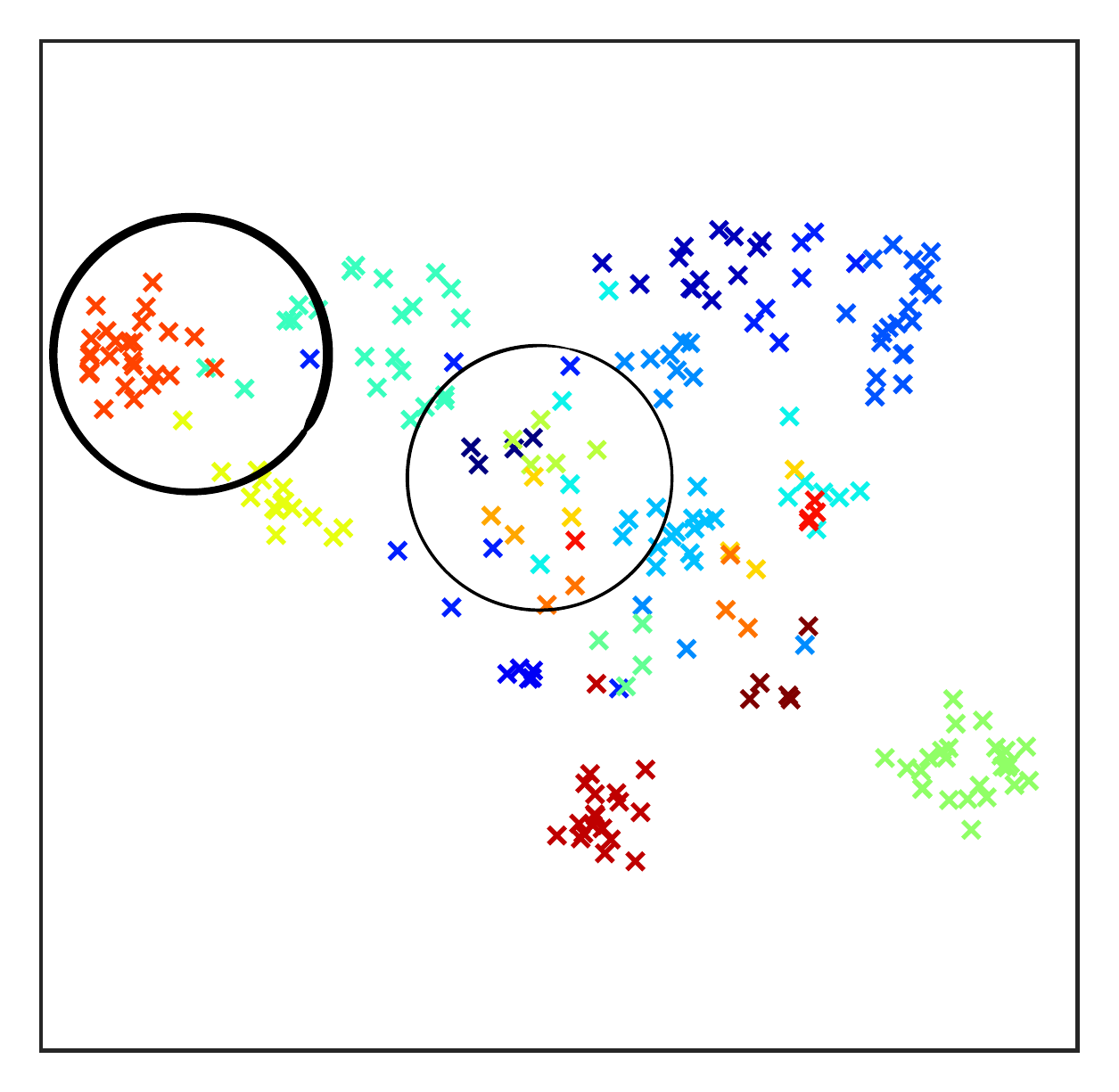}
\caption{\label{fig:yupva}}
\end{subfigure}
\begin{subfigure}[b]{0.495\linewidth}
\includegraphics[trim=0 0 0 0, clip=true,height=6cm]{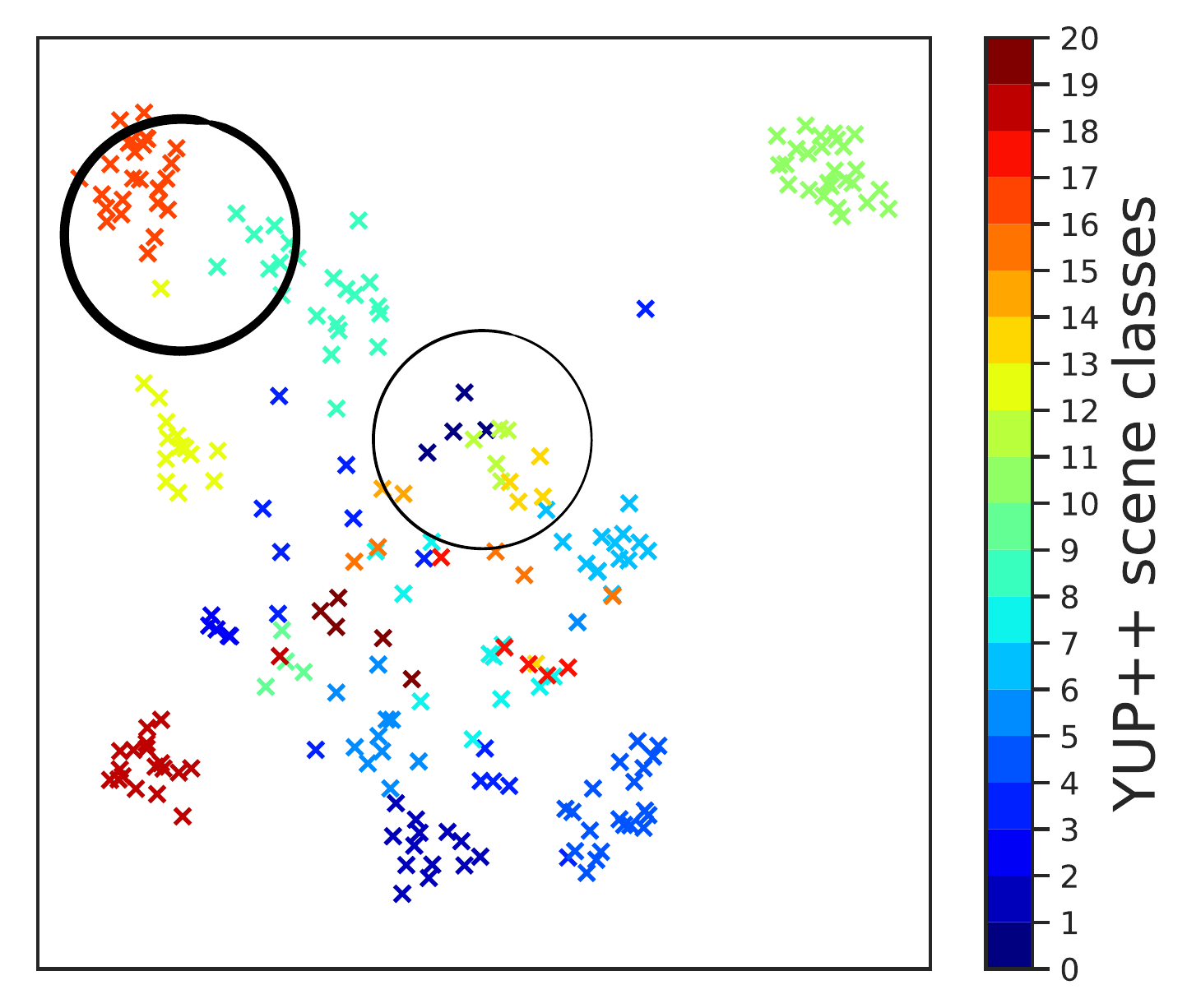}
\caption{\label{fig:yupvb}}
\end{subfigure}
\caption{Visualization of the feature space (from PredNet) for DEEP-HAL  in Fig. \ref{fig:yupva} and DEEP-HAL+ODF  in Fig. \ref{fig:yupvb}  on the YUP++ dataset. For comparison, we circle regions with interesting changes.
}
\vspace{-0.2cm}
\label{fig:yupill}
\end{figure*}

\begin{figure*}[!th]
\centering
\vspace{0.1cm}
\begin{subfigure}[b]{0.48\linewidth}
\includegraphics[trim=0 0 0 0, clip=true,height=6cm]{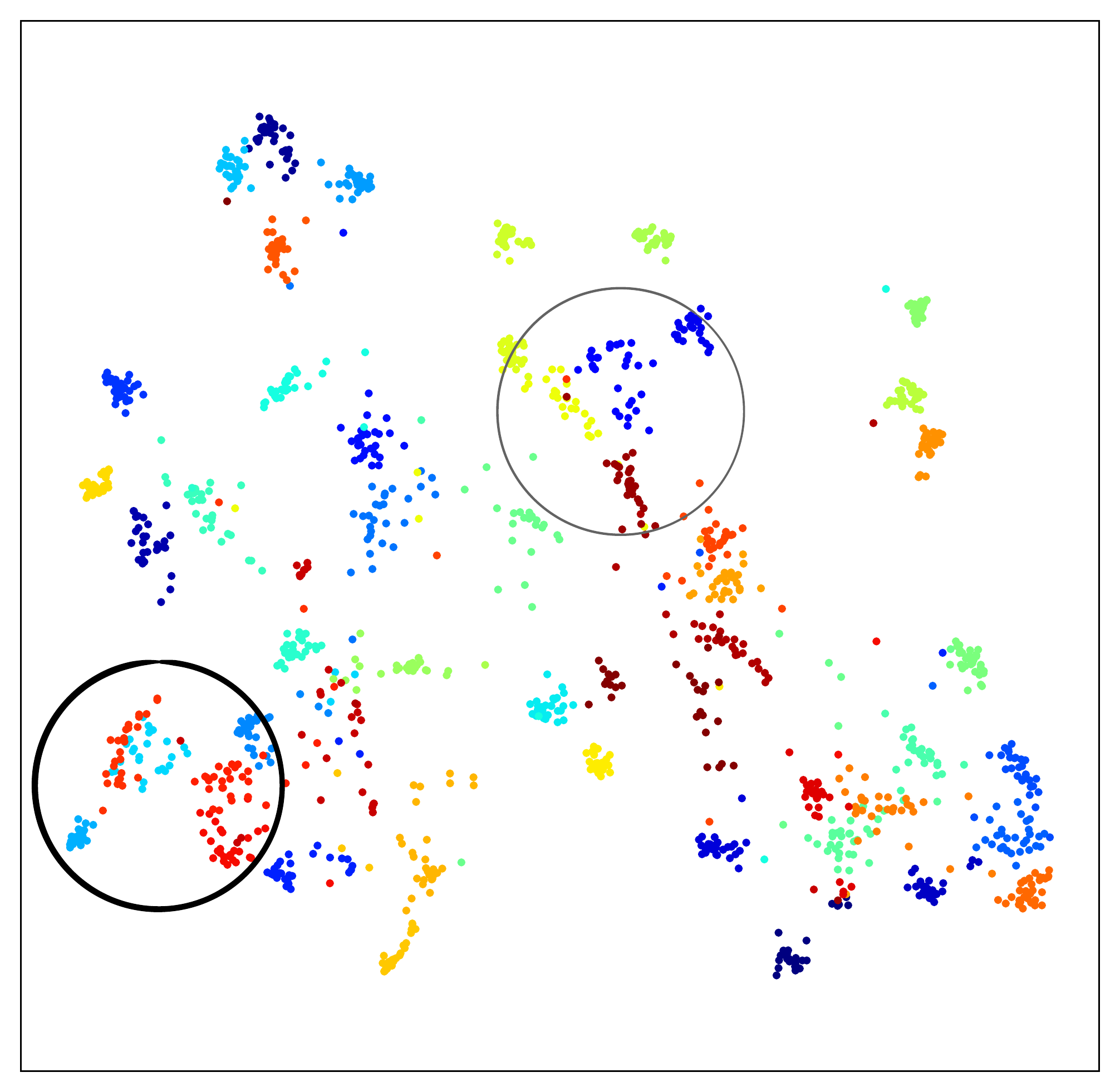}
\caption{\label{fig:hmdb51va}}
\end{subfigure}
\begin{subfigure}[b]{0.495\linewidth}
\includegraphics[trim=0 0 0 0, clip=true,height=6cm]{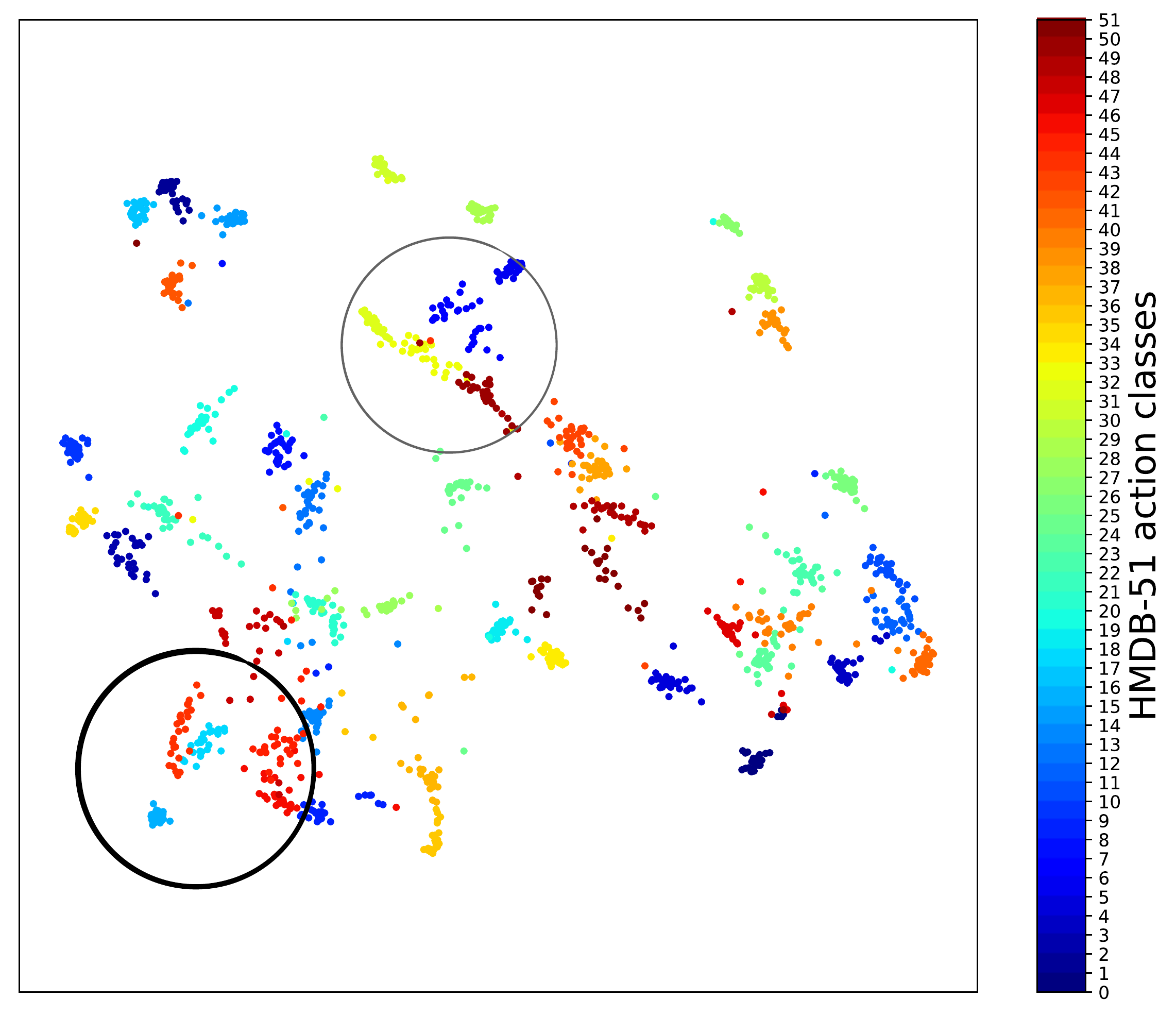}
\caption{\label{fig:hmdb51vb}}
\end{subfigure}
\caption{Visualization of the feature space (from PredNet) for DEEP-HAL in Fig. \ref{fig:hmdb51va} and  DEEP-HAL+ODF in Fig. \ref{fig:hmdb51vb} on the HMDB-51 dataset. For comparison, we circle regions with interesting changes.
}
\vspace{-0.2cm}
\label{fig:hmdb51ill}
\end{figure*}

Table \ref{tab:stat1} introduces timing for object detectors used by our ODF descriptors during training. We note that detections with all four object detectors which we use take $\sim\!1.47$ second per frame. Thus, obtaining four ODF descriptors per clip (uniquely annotated sequence to train or classify) takes between 136 and 441 seconds. Table \ref{tab:stat2} introduces timing for saliency detectors used in our SDF descriptors  during training.  We note that detections with both saliency  detectors which we use take $\sim\!0.9$ second per frame, and obtaining both SDF descriptors per clip takes between 84 and 271 seconds. We do note that the major computational cost is incurred due to detectors rather than our ODF and SDF descriptors proposed in the main paper (their cost is minimal). We further note that the idea of learning these costly representations during training is very valuable. While the total computations per training clip vary between 220 and 712 seconds, during testing time we obtain these representations for free (milliseconds) thanks to DET1,$\cdots$,DET4 and SAL1/SAL2 units from Figure 2 (the main submission). Assuming 25\% of clips in charades for testing, that results in 137 days of computational savings on a single GPU (conversely, 1 day savings on 137 GPUs). Given the obtained 6\% boost on Charades over the baseline without ODF and SDF, and the computational savings, we believe these statistics highlight the value of our approach.

\section{Visualization using UMAP}

\begin{figure}[t]
\hspace{-0.2cm}%
\centering
\vspace{-0.3cm}
\includegraphics[trim=0 0 0 0, clip=true,height=7cm]{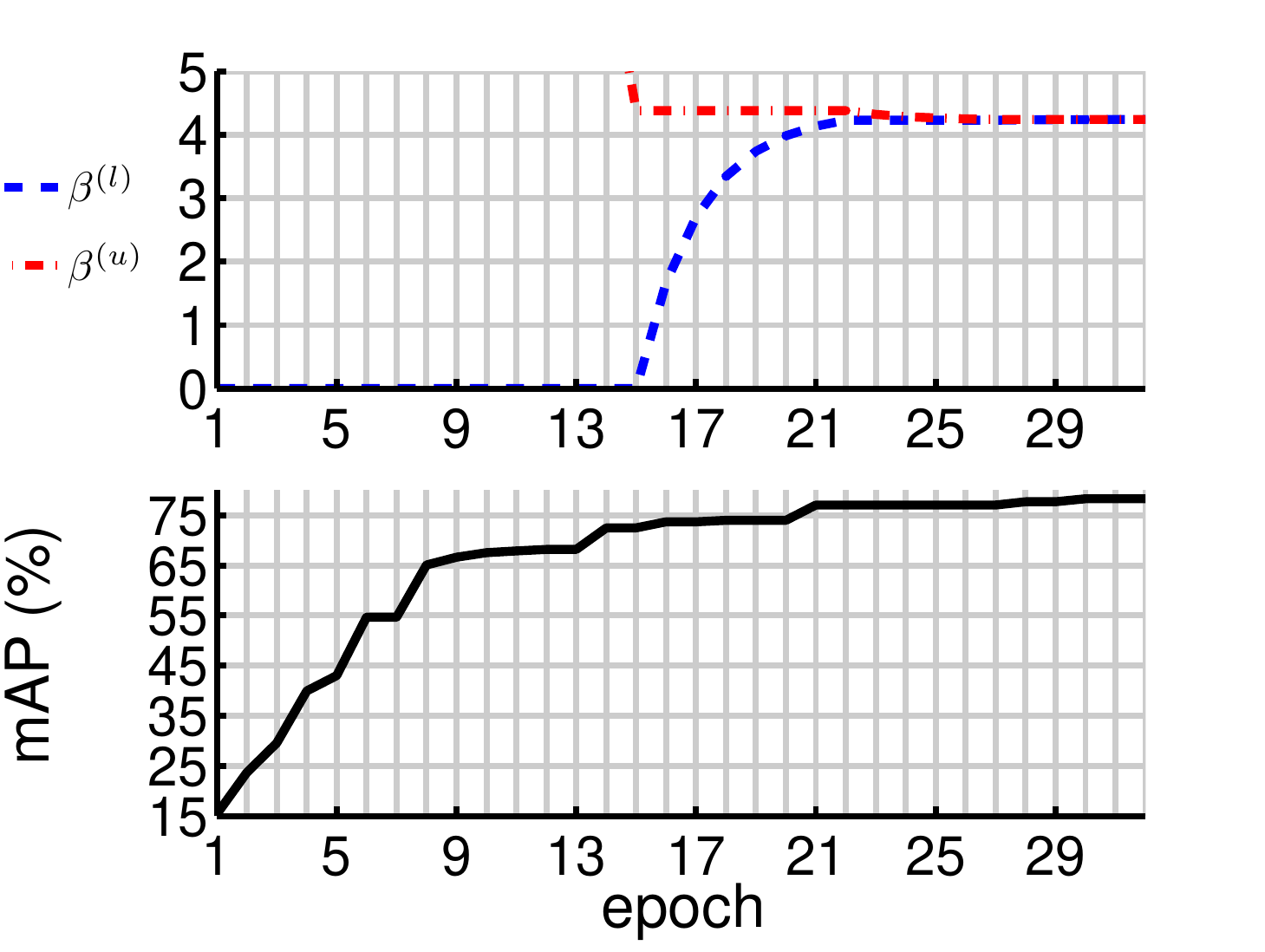}
\caption{Visualization of the Golden-search for the weighting mechanism (final level weighting). ({\em top}) Illustration of how the lower and upper estimates $\beta^{(l)}$ and $\beta^{(u)}$ converge as epochs progress. ({\em bottom}) For every epoch, we set $\beta\!=\!0.5(\beta^{(l)}\!+\!\beta^{(u)})$ and obtain the corresponding validation score (mAP) on MPII ({\em split1}). As the epoch number advances, mAP improves and remains stable as the Golden-search algorithm converges.
}
\vspace{-0.3cm}
\label{fig:gold}
\end{figure}

Figure \ref{fig:yupill} is a visualization performed with UMAP \cite{umap} on the YUP++ dataset. In Fig. \ref{fig:yupva}, top left corner contains samples from classes in red, green, and blue colors which partially overlap. In Fig. \ref{fig:yupvb}, top left corner contains the samples from the corresponding classes in red, green, and blue colors. This time, the samples of these three classes are well separated from each other.

Figure \ref{fig:hmdb51ill} is a visualization performed with UMAP \cite{umap} on the HMDB-51 dataset. In Fig. \ref{fig:hmdb51va}, bottom left corner contains samples from classes in red and blue colors which partially overlap. In Fig. \ref{fig:hmdb51vb}, bottom left corner contains the samples from the corresponding classes in red and blue colors. This time, the class-wise clusters seem to be more clearly delineated and samples of these classes are separated better from each other.

\bibliography{odf-sdf}
\bibliographystyle{plain}

\end{document}